\documentclass[review,authoryear]{elsarticle}
\usepackage[pagewise, modulo,displaymath, mathlines]{lineno}
\usepackage{amsmath}
\usepackage{graphicx}
\usepackage{threeparttable}
\usepackage{multirow}
\usepackage{colortbl}
\usepackage{subfig}
\usepackage{amssymb}
\usepackage{cite}
\usepackage{color}
\journal{Neuroimage}

\begin{document}

\begin{frontmatter}

\title{MRI Cross-Modality \\ NeuroImage-to-NeuroImage Translation}

\author[buaa]{Qianye Yang\corref{equ}}
\ead{QianyeYang@buaa.edu.cn}
\author[buaa]{Nannan Li\corref{equ}}
\ead{linannan0614@foxmail.com}
\author[buaa2]{Zixu Zhao\corref{equ}}
\ead{zixuzhao1218@gmail.com}
\author[cd]{Xingyu Fan\corref{equ}}
\ead{xingyu.fan02@gmail.com}
\author[msra]{Eric I-Chao Chang}
\ead{echang@microsoft.com}
\author[buaa,msra]{Yan Xu\corref{cor}}

\cortext[equ]{These four authors contribute equally to the study}
\cortext[cor]{Corresponding author}
\ead{xuyan04@gmail.com}

\address[buaa]{Research Institute of Beihang University in Shenzhen and State Key Laboratory of Software Development Environment and Key Laboratory of Biomechanics and Mechanobiology of Ministry of Education, Beijing Advanced Innovation Center for Biomedical Engineering, Beihang University, Beijing 100191, China}
\address[buaa2]{School of Electronic and Information Engineering, Beihang University, Beijing 100191, China}
\address[cd]{Bioengineering College of Chongqing University, Chongqing 400044, China}
\address[msra]{Microsoft Research, Beijing 100080, China}

\begin{abstract}
We present a cross-modality generation framework that learns to generate translated modalities from given modalities in MR images without real acquisition. Our proposed method performs NeuroImage-to-NeuroImage translation (abbreviated as N2N) by means of a deep learning model that leverages conditional generative adversarial networks (cGANs). Our framework jointly exploits the low-level features (pixel-wise information) and high-level representations (e.g. brain tumors, brain structure like gray matter, etc.) between cross modalities which are important for resolving the challenging complexity in brain structures. Our framework can serve as an auxiliary method in clinical diagnosis and has great application potential. Based on our proposed framework, we first propose a method for cross-modality registration by fusing the deformation fields to adopt the cross-modality information from translated modalities. Second, we propose an approach for MRI segmentation, translated multichannel segmentation (TMS), where given modalities, along with translated modalities, are segmented by fully convolutional networks (FCN) in a multichannel manner. Both of these two methods successfully adopt the cross-modality information to improve the performance without adding any extra data. Experiments demonstrate that our proposed framework advances the state-of-the-art on five brain MRI datasets. We also observe encouraging results in cross-modality registration and segmentation on some widely adopted brain datasets. Overall, our work can serve as an auxiliary method in clinical diagnosis and be applied to various tasks in medical fields.

\end{abstract}

\end{frontmatter}


\section{Introduction}

\begin{figure}[!htp]
\centering
\includegraphics[width=0.8\textwidth]{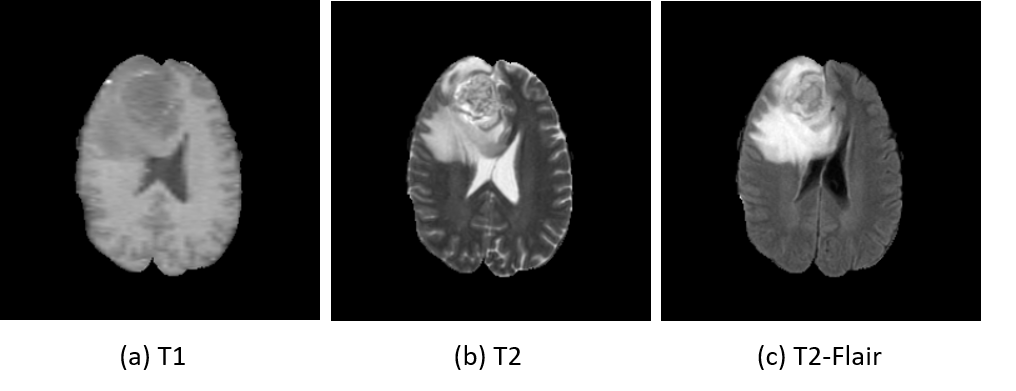}
\caption{Examples of three different modalities: (a) T1, (b) T2, and (c) T2-Flair.}
\label{fig:introduction}
\end{figure}

Magnetic Resonance Imaging (MRI) has become prominent among various medical imaging techniques due to its safety and information abundance. They are broadly applied to clinical treatment for diagnostic and therapeutic purposes. There are different modalities in MR images, each of which captures certain characteristics of the underlying anatomy. All these modalities differ in contrast and function.
Three modalities of MR images are commonly referenced for clinical diagnosis: T1 (spin-lattice relaxation), T2 (spin-spin relaxation), and T2-Flair (fluid attenuation inversion recovery) \citep{Tseng2017Joint}.
T1 images are favorable for observing structures, e.g. gray matter and white matter in the brain; T2 images are utilized for locating tumors; T2-Flair images present the location of lesions with water suppression. Each modality provides a unique view of intrinsic MR parameters. Examples of these three modalities are shown in Fig.\ref{fig:introduction}. Taking full consideration of all these modalities is conducive to MR image analysis and diagnosis.

However, the existence of complete multi-modality MR images is limited by the following factors:
(1)During the scanning process, the imaging of a certain modality usually fails.
(2) Motion artifacts are produced along with MR images. These artifacts are attributed to the difficulty of staying still for patients during scanning (e.g. pediatric population \citep{Rzedzian1983Real}), or motion-sensitive applications such as diffusion imaging \citep{Tsao2010Ultrafast}.
(3) The mapping from one modality to another is hard to learn.
Each of modality captures different characteristics of the underlying anatomy, and the relationship between any two modalities is highly non-linear. Owing to differences in the image characteristics across modalities, existing approaches cannot achieve satisfactory results for cross-modality synthesis as mentioned in \citep{Vemulapalli2016Unsupervised}. For example, when dealing with the paired MRI data, the regression-based approach \citep{Jog2013Magnetic} even lose some information of brain structures.
Synthesizing a translated modality from a given modality without real acquisitions, also known as cross-modality generation, is a nontrivial problem worthy of being studied. Take the transition from T1 (given modality) to T2 (target modality) as an example, $\widehat {T}2$ (translated modality) can be generated through a cross-modality generation framework. In this paper, $\widehat {}$ denotes translated modalities. Cross-modality generation tasks refer to transitions such as from T1 to T2, from T1 to T2-Flair, from T2 to T2-Flair, and vice versa.

Recently, image-to-image translation networks have provided a generic solution for image prediction problems in natural scenes, like mapping images to edges \citep{Xie2015Holistically,Lee2014Deeply}, segments \citep{xu2017gland}, semantic labels \citep{long2015fully} (many to one), and mapping labels to realistic images (one to many). It requires an automatic learning process for loss functions to make the output indistinguishable from reality. The recently proposed Generative Adversarial Network (GAN) \citep{ goodfellow2014generative,pathak2016context,isola2016image,zhang2016stackgan} makes it possible to learn a loss adapting to the data and be applied to multiple translation tasks. Isola et al. \citep{isola2016image} demonstrate that the conditional GAN (cGAN) is suitable for image-to-image translation tasks, where they condition on input images.

\begin{figure}[!htp]
\centering
\includegraphics[width=0.8\textwidth]{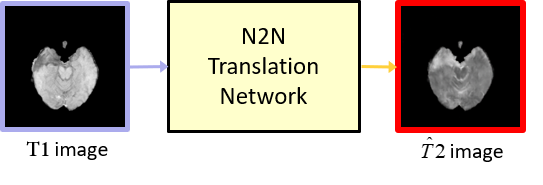}
\caption{Overview of our N2N translation network. It learns to generate translated modality images ($\widehat {T}2$) from given modality images (T1). The red box indicates our translated images.}
\label{fig:intro_gen}
\end{figure}

Previous work on image-to-image translation networks focuses on natural scenes \citep{isola2016image,Tu2007Learning,Lazarow2017Introspective,Jin2017Introspective}, however, such networks' effectiveness in providing a solution for translation tasks in medical scenes remains inconclusive. Motivated by \citep{isola2016image}, we introduce NeuroImage-to-NeuroImage translation networks (N2N) to brain MRI cross-modality generation (see Fig.\ref{fig:intro_gen}). Unlike some classic regression-based approaches that leverage an L1 loss to capture the low-level information, we adopt cGANs to capture high-level information and an L1 loss to ensure low-level information at the same time, which allows us to recover more details from the given modality and reduce the noise generated along with the translated modality.

In this paper, we mainly focus on developing a cross-modality generation framework which provides us with novel approaches of cross-modality registration and segmentation. Our proposed cross-modality generation framework can serve as an auxiliary method in clinical diagnosis and also has great application potential, such as multimodal registration \citep{Roy2013Magnetic}, segmentation \citep{Iglesias2013Is}, and virtual enhancement \citep{Vemulapalli2016Unsupervised}. Among all these applications, we choose cross-modality registration and segmentation as two examples to illustrate the effectiveness of our cross-modality generation framework.

The first application of our proposed framework is cross-modality image registration which is necessary for medical image processing and analysis. With regard to brain registration, accurate alignment of the brain structures such as hippocampus, gray matter, and white matter are crucial for monitoring brain disease like Alzheimer Disease (AD). The accurate delineation of brain structures in MR images can provide neuroscientists with volumetric and structural information on the structures, which has been already achieved by existing atlas-based registrations \citep{Roy2013Magnetic,Eugenio2013A}. However, few of them adopt the cross-modality information from multiple modalities, especially from translated modalities.

\begin{figure}[!htp]
\centering
\includegraphics[width=0.8\textwidth]{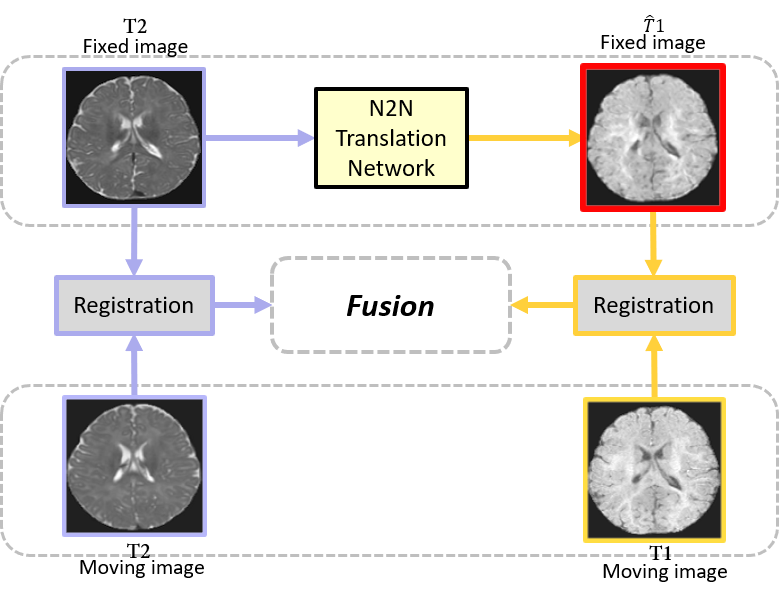}
\caption{Overview of our approach for cross-modality registration. Inputting a given-modality image (T2) to N2N framework yields a translated modality ($\hat{T}1$). Then T2 (moving) is registered to T2 (fixed), T1 (moving) is registered to $\hat{T}1$ (fixed). The deformation generated in the registration process are finally combined in a weighted fusion process, obtaining our final registration result. The red box indicates our translated images.}
\label{fig:intro_reg}
\end{figure}

\begin{figure}[!htp]
\centering
\includegraphics[width=0.8\textwidth]{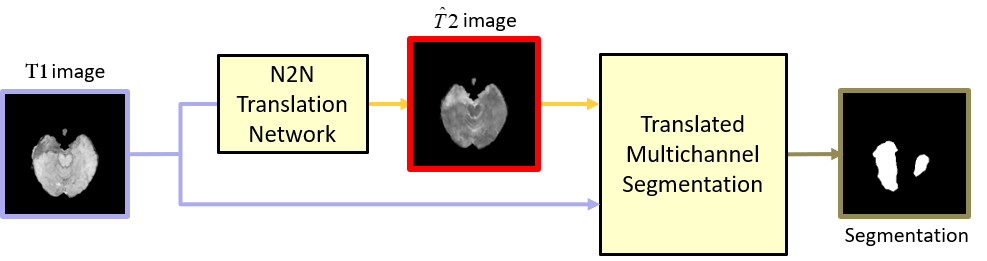}
\caption{Overview of our approach for cross-modality segmentation. First, we input a given-modality image to our N2N translation network to generate a translated-modality image. For instance, given a T1 image, $\widehat {T}2$ images can be generated with our method. Second, the translated modality ($\widehat {T}2$) and its corresponding given modality (T1) are fed into fully convolutional networks (FCN) \citep{long2015fully} for brain segmentation. The red box indicates our translated images.}
\label{fig:intro_seg}
\end{figure}

Here, we propose a new method for cross-modality registration by adopting cross-modality information from our translated modalities. The flowchart is illustrated in Fig.\ref{fig:intro_reg}.
In our method, inputting a given-modality image (e.g. T2 image) to our proposed framework yields a translated modality (e.g. $\hat{T}1$ image). Both two modalities compose our fixed images space (T2 and $\hat{T}1$ images). The moving images including T2 and T1 images are then registered to the identical modality in the fixed images space with a registration algorithm. Specifically, T2 (moving) is registered to T2 (fixed), T1 (moving) is registered to $\hat{T}1$ (fixed). The deformation generated in the registration process are finally combined in a weighted fusion process and then propagate the moving images labels to the fixed images space. It is feasible since the introduction of translated modality provides us with richer anatomical information in comparison with only one modality is given, leading to more precise registration results. Our method is applicable to dealing with cross-modality registration problems by making the most of cross-modality information without adding any extra data at the same time.
The second application of our proposed framework is brain segmentation for MRI data, which also plays an important role in clinical auxiliary diagnosis. However, it is a difficult task owing to the artifacts and in-homogeneities introduced during the real image acquisition \citep{Balafar2010Review,Sasirekha2015Improved}.
To this point, we propose a novel approach for brain segmentation, called translated multichannel segmentation (TMS).
In TMS, as illustrated in Fig.\ref{fig:intro_seg}, the translated modality and its corresponding given modality are fed into fully convolutional networks (FCN) \citep{long2015fully} for brain segmentation. Here, we fine tune Imagenet-FCN model using our MRI images.
Thus we follow its original three-channel network, inputting one translated modality and two given modality images to serve as three channels.
TMS is an effective method for brain segmentation by adding cross-modality information from translated modalities since different MRI modalities have unique tissue contrast profiles and therefore provide complementary information that could be of use to the segmentation process. 
For instance, TMS can improve tumor segmentation performance by adding cross-modality information from translated T2 modality into original T1 modality.

\textbf{Contributions:}
(1) We introduce end-to-end NeuroImage-to-NeuroImage translation networks for cross-modality MRI generation to synthesize translated modalities from given modalities. Our N2N framework can cope with a great many MRI translation tasks using the same objective and architecture.
(2) Registration: We leverage our N2N framework to augment the fixed images space with translated modalities for atlas-based registration. Registering moving images to fixed images and weighted fusion process enable us to make the most of cross-modality information without adding any extra data.
(3) Segmentation: Our proposed approach, translated multichannel segmentation (TMS), performs cross-modality image segmentation by means of FCNs. We input two identical given modalities and one corresponding translated modality into separate channels, which allows us to adopt and fuse cross-modality information without using any extra data.
(4) We demonstrate the universality of N2N framework for cross-modality generation on five publicly available brain datasets. Experiments conducted on two sets of datasets also verify the effectiveness of two applications of our proposed framework. We finally observe competitive generation results of our proposed framework.

\section{Related work}

In this section, we mainly focus on methods related to cross-modality image generation, its corresponding registration and segmentation.

\subsection{Image generation}
Related work on image generation can be broadly divided into three categories: cross-modality synthesis, GANs in natural scenes, and GANs in medical images.

\textbf{Cross-modality synthesis: }In order to synthesize one modality from another, a rich body of algorithms have been proposed using non-parametric methods like nearest neighbor (NN) search \citep{Freeman2000Learning}, random forests \citep{Jog2013Magnetic}, coupled dictionary learning \citep{Roy2013Magnetic}, and convolutional neural network (CNN) \citep{Nguyen2015Cross}, etc. They can be broadly categorized into two classes: \textbf{(1) Traditional methods.} One of the classical approaches is an atlas-based method proposed by Miller et al. \citep{Miller1993Mathematical}. The atlas contains pairs of images with different tissue contrasts co-registered and sampled on the same voxel locations in space. An example-based approach is proposed to pick several NNs with similar properties from low-resolution images to generate high-resolution brain MR images using a Markov random field \citep{Rousseau2008Brain}. In \citep{Jog2013Magnetic}, a regression-based approach is presented where a regression forest is trained using paired data from a given modality to a target modality. Later, the regression forest is utilized to regress target-modality patches from given modality patches. \textbf{(2) Deep learning based methods.} Nguyen et al. \citep{Nguyen2015Cross} present a location-sensitive deep network (LSDN) to incorporate spatial location and image intensity feature in a principled manner for cross-modality generation. Vemulapalli et al. \citep{Vemulapalli2016Unsupervised} propose a general unsupervised cross-modal medical image synthesis approach that works without paired training data. Huang et al. \citep{Huang2017Simultaneous} attempt to jointly solve the super-resolution and cross-modality generation problems in 3D medical imaging using weakly-supervised joint convolutional sparse coding.

Our image generation task is essentially similar to these issues. We mainly focus on developing a novel and simple framework for cross-modality image generation and we choose paired  MRI data as our case rather than unpaired data to improve the performance. To this point, we try to develop a 2D framework for cross-modality generation tasks according to 2D MRI  principle. The deep learning based methods \citep{Vemulapalli2016Unsupervised,Huang2017Simultaneous} are not perfectly suitable for our case on the premise of our paired data and MRI principle. We thus select the regression-based approach \citep{Jog2013Magnetic} as our baseline.

\textbf{GANs in natural scenes: }
Recently, a Generative Adversarial Network (GAN) has been proposed by Goodfellow et al. \citep{goodfellow2014generative}. They adopt the concept of a min-max optimization game and provide a thread to image generation in unsupervised representation learning settings. To conquer the immanent hardness of convergence, Radford et al. \citep{radford2015unsupervised} present a deep convolutional Generative Adversarial Network (DCGAN). However, there is no control of image synthesis owing to the unsupervised nature of unconditional GANs. Mirza et al. \citep{mirza2014conditional} incorporate additional information to guide the process of image synthesis. It shows great stability refinement of the model and descriptive ability augmentation of the generator. Various GAN-family applications have come out along with the development of GANs, such as image inpainting \citep{pathak2016context}, image prediction \citep{isola2016image}, text-to-image translation \citep{zhang2016stackgan} and so on. Whereas, all of these models are designed separately for specific applications due to their intrinsic disparities.
To this point, Isola et al. \citep{isola2016image} present a generalized solution to image-to-image translations in natural scenes. Our cross-modality image generation is inspired by \citep{isola2016image} but we focus on medical images generation as opposed to natural scenes.

\textbf{GANs in medical images: }
In spite of the success of existing approaches in natural scenes, there are few applications of GANs to medical images. Nie et al. \citep{nie2016medical} estimate CT images from MR images with a Context-Aware GAN model. Wolterink et al. \citep{WolterinkGenerative} demonstrate that GANs are applicable to transforming low-dose CT into routine-dose CT images. However, all these methods are designed for specific rather than general applications. Loss functions need to be modified when it comes to multi-modality transitions. Thus, a general-purpose strategy for medical modality transitions is of great significance. Fortunately, this is achieved by our N2N cross-modality image generation framework.

\subsection{Image registration}
A successful image registration application requires several components that are correctly combined, like the cost function and the transformation model.
The cost function, also called similarity metrics, measures how well two images are matched after transformation. It is selected with regards to the types of objects to be registered. As for cross-modality registration, commonly adopted cost functions are mutual information (MI) \citep{Viola1997Alignment} and cross-correlation (CC) \citep{Penney1998A}.
Transformation models are determined according to the complexity of deformations that need to be recovered. Some common parametric transformation models (such as rigid, affine, and B-Splines transformation) are enough to recover the underlying deformations \citep{Rueckert1999Nonrigid}.

Several image registration toolkits such as ANTs \citep{Avants2009Advanced} and Elastix \citep{Klein2010elastix} have been developed to facilitate research reproduction. These toolkits have effectively combined commonly adopted cost functions and parametric transformation models. They can estimate the optimal transformation parameters or deformation fields based on an iterative framework. In this work, we choose ANTs and Elastix to realize our cross-modality registration. More registration algorithms can be applied to our method.

\subsection{Image segmentation}
A rich body of image segmentation algorithms exists in computer vision \citep{Pinheiro2015From,dou2016automatic,long2015fully,xu2017gland}. We discuss two that are closely related to our work.

The Fully Convolutional Network (FCN) proposed by Long et al. \citep{long2015fully} is a semantic segmentation algorithm. It is an end-to-end and pixel-to-pixel learning system which can predict dense outputs from arbitrary-sized inputs. Inspired by \citep{long2015fully}, TMS adopts similar FCN architectures but focuses on fusing information of different modalities in a multichannel manner.

Xu et al. \citep{xu2017gland} propose an algorithm for gland instance segmentation, where the concept of multichannel learning is introduced. The proposed algorithm exploits features of edge, region, and location in a multichannel manner to generate instance segmentation. By contrast, TMS leverages features in translated modalities to refine the segmentation performance of given modalities.

\section{MRI Cross-Modality Image Generation}

\begin{figure}[!ht]
\centering
\includegraphics[width=120mm]{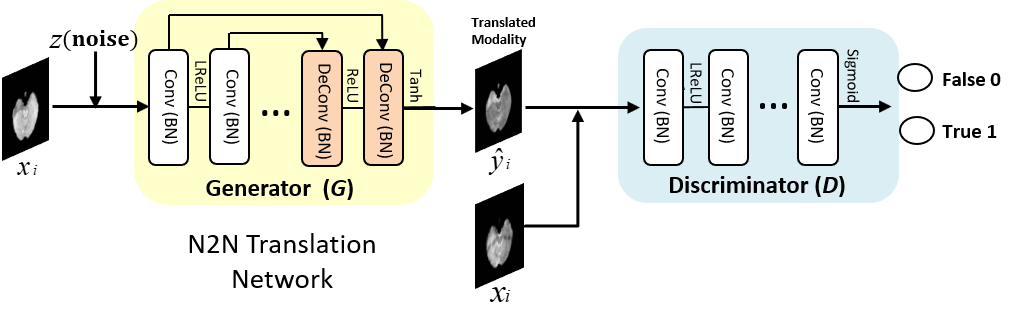}
\caption{Overview of our end-to-end N2N translation network for cross-modality generation. Notice that our training set is denoted as $S=\{(x_i, y_i), i=1, 2, 3, \ldots, n\}$, where $x_i$ and $y_i$ refer to the $i$th input given-modality image and its corresponding target-modality image. The training process involves two aspects. On the one hand, given an input image $x_i$ and a random noise vector $z$, generator $G$ aims to produce indistinguishable images $\hat{y}_i$ from the real images $y_i$. On the other hand, discriminator $D$ evolves to distinguish between translated-modality images $\hat{y}_i$ generated by $G$ and the real images $y_i$. The output of $D$ is 0 or 1, where 0 represents synthesized images and 1 represents the real data. In the generation process, translated-modality images can be synthesized through the optimized $G$. }
\label{fig:generation_flowchart}
\end{figure}

In this section, we mainly learn an end-to-end mapping from given-modality images to target-modality images. We introduce NeuroImage-to-NeuroImage (N2N) translation networks to cross-modality generation. Here, cGANs are used to realize NeuroImage-to-NeuroImage translation networks.
The flowchart of our algorithm is illustrated in Fig.\ref{fig:generation_flowchart}.

\subsection{Training}
We denote our training set as $S=\{(x_i, y_i), i=1, 2, 3, \ldots, n\}$, where $x_i$ refers to the $i$th input given-modality image, and $y_i$ indicates the corresponding target-modality image.
We subsequently drop the subscript $i$ for simplicity, since we consider each image holistically and independently.
Our goal is to learn a mapping from given-modality images $\{x_i\}_{i=1}^{n}\in X$ to target-modality images $\{y_i\}_{i=1}^{n}\in Y$.
Thus, given an input image $x$ and a random noise vector $z$, our method can synthesize the corresponding translated-modality image $\widehat{y}$.
Take the transition from T1 to T2 as an instance.
Similar to a two-player min-max game, the training procedure of GAN mainly involves two aspects:
On one hand, given an input image T1 ($x$), generator $G$ produces a realistic image $\hat{T}2$ ($\hat{y}$) towards the real data T2 ($y$) in order to puzzle discriminator $D$. On the other hand, $D$ evolves to distinguish synthesized images $\hat{T}2$ ($\hat{y}$) generated by $G$ from the real data T2 ($y$). The overall objective function is defined:
\begin{gather}
\mathcal{L}_{cGAN}(G,D) = \mathbb{E}_{x,y\sim p_{data}(x,y)}[\log D(x,y)] + \notag \\
 \mathbb{E}_{x\sim p_{data}(x),z~p_{z}(z)}[\log (1-D(x,G(x,z))],
\end{gather}
where $p_{data}(x)$ and $p_{data}(z)$ refer to the distributions over data $x$ and $z$, respectively.
$G$ is not only required to output realistic images to fool $D$, but also to produce high-quality images close to the real data.
Existing algorithms \citep{pathak2016context} have found it favorable to combine traditional regularization terms with the objective function in GAN.
An L1 loss, as described in \citep{isola2016image}, usually guarantees the correctness of low-level features and encourages less blurring than an L2 loss.
Thus, an L1 loss term is adopted into the objective function in our method. The L1 loss term is defined as follows:
\begin{equation}
\mathcal{L}_{L1}(G) = \mathbb{E}_{x,y\sim p_{data}(x,y),z\sim p_{z}(z)}[\lVert y-G(x,z) \lVert_{1}].
\end{equation}
The overall objective function is then updated to:
\begin{equation}
\mathcal{L} = \mathcal{L}_{cGAN}(G,D) + \lambda\mathcal{L}_{L1}(G),
\label{eq:loss}
\end{equation}
where $\lambda$ is a hyper-parameter specified manually to balance the adversarial loss and L1 loss. The appropriate weight of $\lambda$ is based on the cross-validation of training data. A value of $100$ is eventually selected for $\lambda$.

Following \citep{isola2016image}, the optimization is an iterative training process with two steps: (1) fix parameters of $G$ and optimize $D$; (2) fix parameters of $D$ and optimize $G$. The overall objective function can be formulated as follows:
\begin{equation}
\mathnormal{G}^{*} = arg\;  \min_G \; \max_D \mathcal{L}_{cGAN}(G,D) + \lambda\mathcal{L}_{L1}(G).
\end{equation}

Here, the introduction of $z$ enables it to match any distribution rather than just a delta function. As \citep{srivastava2013improving} described, dropout can also be interpreted as a way of regularizing a neural network by adding noise to its hidden units. Thus we replace the noise vector $z$ with several dropout layers in G to achieve the same effect.

In addition, we also explore the effectiveness of each component in our objective function. 
Generators with different loss functions are defined as follows:
$cGAN$: Generator $G$ together with an adversarial discriminator conditioned on the input;
$L1$: Generator $G$ with an L1 loss. It is essentially equivalent to a traditional CNN architecture with least absolute deviation;
$cGAN+L1$: Generator $G$ with both an L1 loss term and an adversarial discriminator conditioned on the input.

\subsection{Network architecture}

Our cross-modality generation framework is composed of two main submodels, \textbf{generator ($G$)} and \textbf{discriminator ($D$)}. It is similar to traditional GANs \citep{goodfellow2014generative}.

\textbf{Generator.}
Although appearances of input and output images are different, their underlying structures are the same. Shared information (e.g. identical structures) needs to be transformed in the generative network. In this case, encoder-decoder networks with an equal number of down-sampling layers and up-sampling layers are proposed as one effective generative network \citep{johnson2016perceptual,pathak2016context,wang2016generative,yoo2016pixel,zhou2016learning}.
However, it is a time-consuming process when all mutual information between input and output images (such as structures, edges and so on) flows through the entire network layer by layer.
Besides, the network efficiency is limited due to the presence of a bottleneck layer which restricts information flow.
Thus, skip connections are added between mirrored layers in the encoder-decoder network, following the ``U-Net'' shape in \citep{ronneberger2015u}.
These connections speed up information transmission since the bottleneck layer is ignored, and help to learn matching features for corresponding mirrored layers.

The architecture of $G$ has 8 convolutional layers, each of which contains a convolution, a Batch Normalization, and a leaky ReLu activation \citep{Ioffe2015Batch} (a slope of 0.2)  with numbers of filters at 64, 128, 256, 512, 512, 512, 512, and 512 respectively. Following them are 8 deconvolutional stages, each of which includes a deconvolution, a Batch Normalization, and an unleaky ReLu \citep{Ioffe2015Batch} (a slope of 0.2) with numbers of filters at 512, 1024, 1024, 1024, 1024, 512, 256, and 128 respectively. It ends with a tanh activation function.

\begin{figure*}[!ht]
\centering
\includegraphics[width=\textwidth]{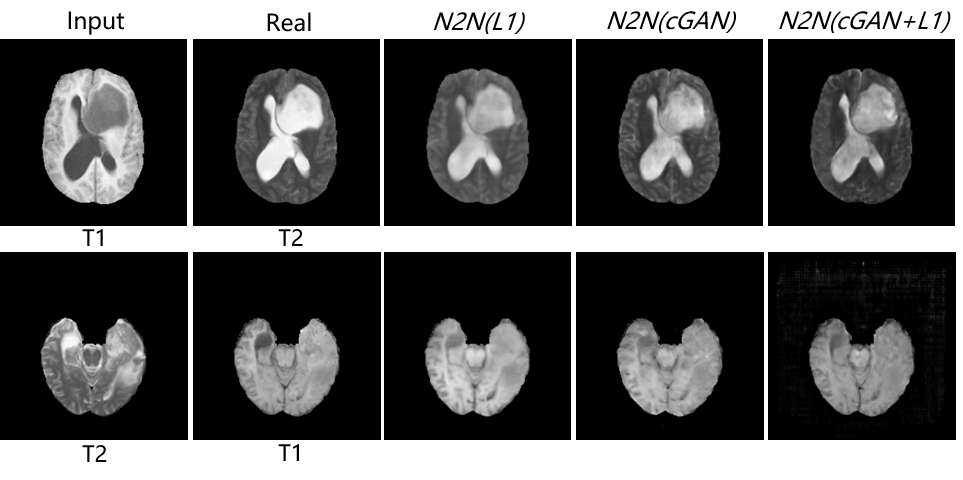}
\caption{Samples of cross-modality generation results on \textit{BraTs2015}. The left two columns respectively show the inputting given-modality images and the real target-modality images. The right three column shows results of N2N framework with different loss functions ($L1, cGAN, cGAN+L1$).}
\label{fig:sample_generation}
\end{figure*}

\textbf{Discriminator.}
GANs can generate images that are not only visually realistic but also quantitatively comparable to the real images. Therefore, an adversarial discriminator architecture is employed to confine the learning process of $G$. $D$ identifies those generated outputs of $G$ as false (label 0) and the real data as true (label 1), then providing feedback to $G$. PixelGANs \citep{isola2016image} have poor performance on spatial sharpness, and ImageGANs \citep{isola2016image} with many parameters are hard to train. In contrast, PatchGANs \citep{isola2016image} enable sharp outputs with fewer parameters and less running time since PatchGANs have no constraints on the size of each patch. We thus adopt a PatchGAN classifier as our discriminator architecture.
Unlike previous formulations \citep{iizuka2016let,larsson2016learning} that regard the output space as unstructured, our discriminator penalizes structures at the scale of image patches. In this way, high-level information can be captured under the restriction of $D$, and low-level information can be ensured by an L1 term.
As shown in Fig.\ref{fig:sample_generation}, training with only the L1 loss gives obscure translated images that lack some discernible details. Under the same experimental setup, the results on the \textit{BraTs2015} dataset are improved notably with the combination of the adversarial loss and L1 loss.

The architecture of $D$ contains four stages of convolution-BatchNorm-ReLu with the kernel size of (4,4).
The numbers of filters are 64, 128, 256, and 512 for convolutional layers. Lastly, a sigmoid function is used to output the confidence probability that the input data comes from real MR images rather than generated images.


\section{Application}
\label{application}
In this section, we choose cross-modality registration and segmentation from multiple applications as two examples to verify the effectiveness of our proposed framework. Details of our approaches and algorithms are discussed in the following subsections.
\subsection{Cross-Modality Registration }
\begin{figure}[!ht]
\centering
\includegraphics[width=120mm]{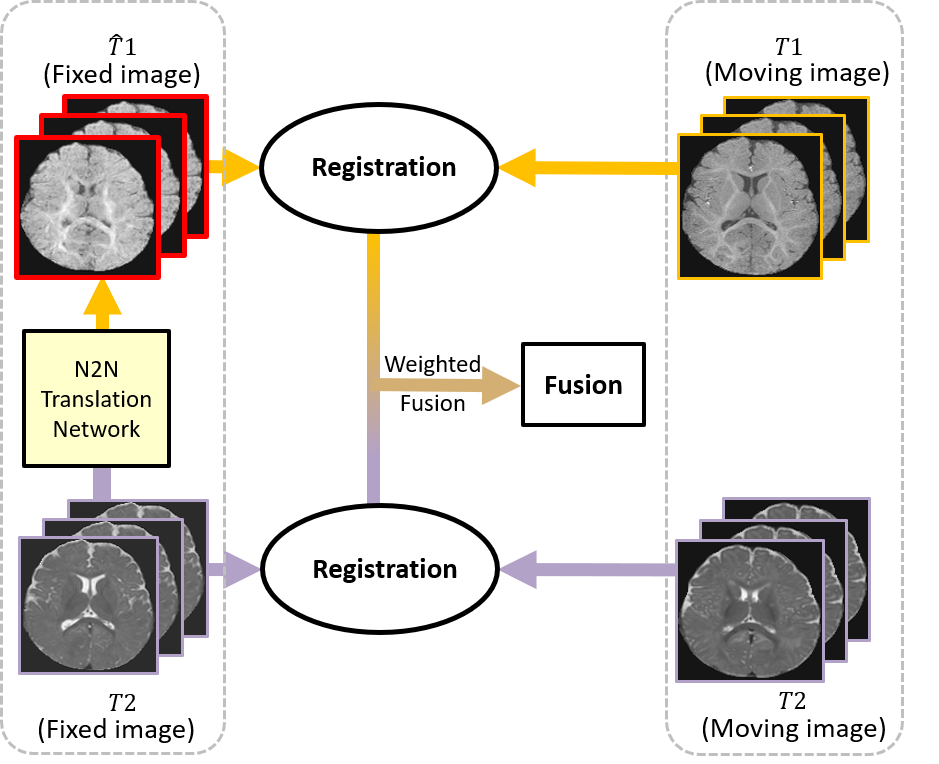}
\caption{Flowchart of our approach for cross-modality registration. In the fixed space, inputting T2 images into N2N framework yields $\hat{T}1$ images.  T2 (moving) images are registered to T2 (fixed) images. T1 (moving) images are registered to $\hat{T}1$ (fixed) images. The corresponding deformations generated after registrations are combined in a weighted fusion process. Then we employ the fused deformation to the segmentation labels of moving images, outputting the registered segmentation labels of fixed images. The red box indicates our translated images.}
\label{fig:registration_flowchart}
\end{figure}
The first application of our cross-modality generation framework is to use the translated modality for cross-modality image registration.
Our method is inspired by an atlas-based registration, where the moving image is registered to the fixed image with a non-linear registration algorithm. Images after registration are called the warped images. Our method contains four steps: (1) We first build our fixed images space with only one modality images being given. We use T1 and T2 images as one example to illustrate our method. Given T2 images, our fixed images space can consist of T2 and $\hat{T}1$ images by using our cross-modality generation framework. The moving images space commonly consists of both T2 and T1 images from $n$ subjects. (2) The second step is to register the moving images to the fixed images, constructing $n$ corresponding atlases. Since multiple atlases encompass richer anatomical variability than a single atlas, we used multi-atlas-based rather than single-atlas-based registration approach. For any fixed subject, we register all $n$ moving images to the fixed images and the deformation field that aligns the moving image with the fixed image can be automatically computed with a registration algorithm. As illustrated in Fig.\ref{fig:registration_flowchart}, T2 images from the moving images space are registered to T2 images from the fixed images space and T1 images from the moving images space are registered to $\hat{T}1$ images from the fixed images space. (3) The deformations generated in (2) are combined in a weighted fusion process, where the cross-modality information can be adopted. We fuse the deformations generated from T2 registrations with deformations generated from $\hat{T}1$ registrations (see Fig.\ref{fig:registration_flowchart}). (4) Applying the deformations to the atlas segmentation labels can yield $n$ registered segmentation labels of fixed images. For any fixed subject, we obtain the final registration results by averaging the $n$ registered labels of the fixed subject.

Among multiple registration algorithms, we select ANTs \citep{Avants2009Advanced} and Elastix \citep{Klein2010elastix} to realize our method. Three stages of cross-modality registration are adopted via ANTs. The first two stages are modeled by rigid and affine transforms with mutual information. In the last stage, we use SyN with local cross-correlation, which is demonstrated to work well with cross-modality scenarios without normalizing the intensities \citep{Boltcheva2009Evaluation}. For Elastix, affine and B-splines transforms are used to model the nonlinear deformations of the atlases. Mutual information is adopted as the cost function.

\subsection{Cross-Modality Segmentation }
\label{segmentation}

\begin{figure}[!htp]
\centering
\includegraphics[width=\textwidth]{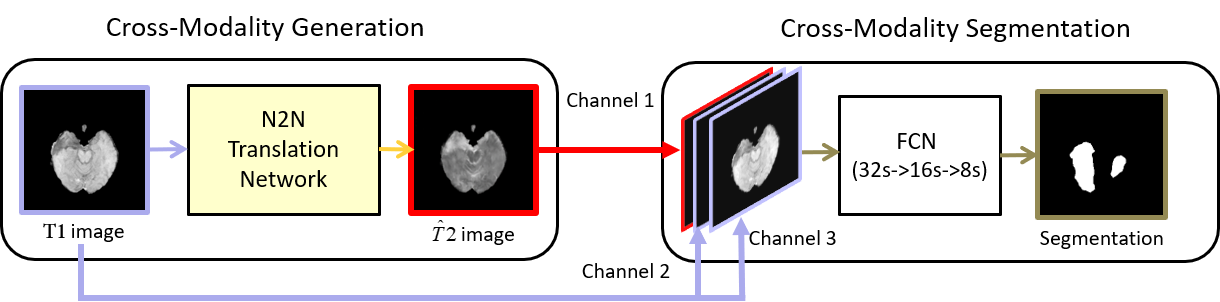}
\caption{Flowchart of our approach for cross-modality segmentation. First, we input a given-modality image to our N2N translation network to generate a translated-modality image. For instance, given a T1 image, $\widehat {T}2$ images can be generated with our method. Second, two identical given-modality images and one corresponding translated-modality image are fed to channels 1, 2, and 3 and segmented by FCN networks. Under the standard FCN-32s, standard FCN-16s, and standard FCN-8s settings, we output our segmentation results. The red box indicates our translated images.}
\label{fig:all-flowchart}
\end{figure}

We propose a new approach for MR image segmentation based on cross-modality images, namely translated multichannel segmentation (TMS).
The main focus of TMS is the introduction of the translated-modality images obtained in our proposed framework, which enriches the cross-modality information without any extra data.
TMS inputs two identical given-modality images and one corresponding translated-modality image into three separate channels which are conventionally used for RGB images. Three input images are then fed into FCN networks for improving segmentation results of given-modality images. Here, we employ the standard FCN-8s \citep{long2015fully} as the CNN architecture of our segmentation framework because it can fuse multi-level information by combining feature maps of the final layer and last two pooling layers.  Fig.\ref{fig:all-flowchart} depicts the flowchart of our segmentation approach.


We denote our training dataset as $S=\{(x_i, \hat{y}_i, l_i), i=1, 2, 3, \ldots, n\}$, where $x_i$ refers to the $i$th given-modality image, $\hat{y}_i$ indicates the $i$th corresponding translated-modality image obtained in our proposed framework, and $l_i$ represents the corresponding segmentation label.
We denote the parameters of the FCN architecture as $\theta$ and the model is trained to seek optimal parameters $\theta^{*}$.
During testing, given an input image $x$, the segmentation output $\hat{l}$ is defined as below:
\begin{equation}
\mathnormal{P}(\hat{l}=k| x; \theta^{*})= s_k(h(x, \theta^{*})),
\end{equation}
where $k$ denotes the total number of classes, $h(\cdot)$ denotes the feature map of the hidden layer, $s(\cdot)$ refers to the softmax function and $s_k$ indicates the output of the $k$th class.

\section{Experiments and results}
\label{Experiments}
In this section, we demonstrate the generalizability of our framework for MR image generation and apply it to cross-modality registration and segmentation. We first conduct a large number of experiments on five publicly available datasets for MR image generation (\textit{BraTs2015, Iseg2017, MRBrain13, ADNI, RIRE}). Then we choose \textit{Iseg2017} and \textit{MRBrain13} for cross-modality registration. We finally choose \textit{BraTs2015} and \textit{Iseg2017} for cross-modality segmentation. Among these five MRI datasets, the \textit{BraTs2015}, \textit{Iseg2017}, and \textit{MRBrain13} datasets provide ground truth segmentation labels.

\subsection{Implementation details}
All our models are trained on NVIDIA Tesla K80 GPUs. Our code\footnote{Implementation details can be found at https://github.com/QianyeYang/MRI-Img2ImgTrans.} will be publicly released upon acceptance.

Generation: We train the models on a torch7 framework \citep{Collobert2011Torch7} using Adam optimizer \citep{Kingma2014Adam} with a momentum term $\beta1=0.5$. The learning rate is set to 0.0002.
The $batchsize$ is set to 1 because our approach can be regarded as ``instance normalization'' when $batchsize=1$ due to the use of batch normalization. As demonstrated in \citep{Ulyanov2016Instance}, instance normalization is effective at generation tasks by removing instance-specific information from the content image.
Other parameters follow the reference \citep{isola2016image}. All experiments use 70$\times$70 PatchGANs.

Registration: A Windows release 2.1.0 version of ANTs \citep{Avants2009Advanced} as well as its auxiliary registration tools are used in our experiments. As for the Elastix \citep{Klein2010elastix}, a Windows 64 bit release 4.8 version is adopted. All the registration experiments are run in a Microsoft High-Performance Computing cluster with 2 Quad-core Xeon 2.43 GHz CPU for each compute node. We choose the parameters by cross-validation. For ANTs, we use the parameters in \citep{Wang2013Multi}. For Elastix, we adopt the parameters in \citep{Artaechevarria2009Combination}.

Segmentation: We implement standard FCN-8s on a publicly available MXNET toolbox \citep{Chen2015MXNet}. A pre-trained VGG-16 model, a trained FCN-32s model, and a trained FCN-16s model are used for initialization of FCN-32s, FCN-16s, and FCN-8s respectively. The learning rate is set to 0.0001, with a momentum of 0.99 and a weight decay of 0.0005. Other parameters are set to the defaults in \citep{long2015fully}.

\subsection{Cross-Modality Generation}

\textbf{Evaluation metrics.}
We report results on mean absolute error (MAE), peak signal-to-noise ratio (PSNR), mutual information (MI),  Structural Similarity Index (SSIM) and FCN-score.

We follow the definition of MAE in \citep{scikit-learn}:
\begin{equation}
\mathnormal{MAE} = \frac{1}{256\times256} \sum_{i=0}^{255} \sum_{j=0}^{255} \| \hat{y}(i,j)-y(i,j)\|,
\end{equation}
where target-modality image $y$ and translated-modality image $\hat{y}$ both have a size of $256\times 256$ pixels, and $(i,j)$ indicates the location of pixels.

PSNR\citep{Hore2010Image} is defined as below:
\begin{equation}
\mathnormal{PSNR} = 10\log{10}{\frac{MAX^{2}}{MSE}},
\end{equation}
where MAX is the maximum pixel value of two images and MSE is the mean square error between two images.

MI is used as a cross-modality similarity measure \citep{Pluim2003Mutual}. It is robust to variations in modalities and calculated as:
\begin{equation}
I(y;\hat{y}) = \sum_{m \in y} \sum_{n \in \hat{y}} p(m,n)\log \left( \frac{p(m,n)}{p(m)p(n)} \right),
\end{equation}
where $m,n$ are the intensities in target-modality image $y$ and translated-modality image $\hat{y}$ respectively. $p(m,n)$ is the joint probability density of $y$ and $\hat{y}$, while $p(m)$ and $p(n)$ are marginal densities.

SSIM \citep{Wang2009Mean} is defined as follows:
\begin{equation}
\mathnormal{SSIM(x,y)} = \frac{(2\mu_{x}\mu_{y}+c_{1})(2\sigma_{xy}+c_{2})}{({\mu}_x^2+{\mu}_y^2+c_1)({\sigma}_x^2+{\sigma}_y^2+c_2)},
\end{equation}
where $\mu_{x}$ and $\mu_{y}$ denote the mean values of original and distorted images. ${\sigma}_x$ and ${\sigma}_y$ denote the standard deviation of original and distorted images, and ${\sigma}_{xy}$ is the covariance of both images.

FCN-score is used to capture the joint statistics of data and evaluate synthesized images across the board. It includes accuracy and Dice. On one hand, accuracy consists of the mean accuracy of all pixels (denoted as ``all'' in the tables) and per-class accuracy (such as mean accuracy of tumors, gray matter, white matter, etc.). On the other hand, the Dice is defined as follows: $(2 |H\cap G|)/(|H|+|G|)$ where $G$ is the ground truth map and $H$ is the prediction map.

Here, we follow the definitions of FCN-score in \citep{isola2016image} and adopt a pre-trained FCN to evaluate our experiment results. The semantic segmentation task in essence is to label each pixel with its enclosing object or region class. Pre-trained semantic classifiers are used to measure the discriminability of the synthesized images as a fake-metric. If synthesized images are plausible, classifiers pre-trained on real images would classify synthesized images correctly as well. Take the transition from T1 to T2 for instance.
T2 images (training data) are utilized to fine tune an FCN-8s model. Both T2 (test data/real data) and $\widehat {T}2$ (synthesized data) images are subsequently segmented through the well-trained model. We score the segmentation (classification) accuracy of synthesized images against the real images.
The gap of FCN-score between T2 images and $\widehat {T}2$ images quantitatively evaluates the quality of $\widehat {T}2$ images.

\textbf{Datasets.}
The data preprocessing mainly contains three steps. (1) Label Generation: Labels of necrosis, edema, non-enhancing tumor, and enhancing tumor are merged into one label, collectively referred to as tumors. Labels of Grey Matter (gm) and White Matter (wm) remain the same. Thus, three types of labels are used for training: tumors, gm, and wm. (2) Dimension Reduction: We slice the original volumetric MRI data along the z-axis because our network currently only supports 2D input images. For example, the 3D data from BraTs2015 datasets, with a size of $240\times240\times155$ voxels (respectively representing the pixels of x-, y-, z-direction), is sliced to 2D data ($155\times220$, 155 slices and 220 subjects). (3) Image Resizing and Scaling: All 2D images are then resized to a resolution of $256\times256$ pixels, after which we generate the 2D input images. Then the input images are scaled from [0, 255] to [0.0, 1.0] and normalized with mean value of 0.5 and standard deviation of 0.5. So, all the input data are normalized in range [-1.0, 1.0].
Note that different modalities of the same subject from five brain MRI datasets that we choose are almost voxel-wise spatially aligned. We do not choose to coregister the data in our datasets since this is beyond the scope of our discussion. We respectively illustrate five publicly available datasets used for cross-modality MRI generation.

(1)\emph{BraTs2015}: The BraTs2015 dataset \citep{Menze2015The} contains multi-contrast MR images from 220 subjects with high-grade glioma, including T1, T2, T2-Flair images and corresponding labels of tumors.
We randomly select 176 subjects for training and the rest for testing. 1924 training images are trained for 600 epochs with batch size 1. 451 images are used for testing.

(2)\emph{Iseg2017}: The Iseg2017 dataset \citep{wang2015links} contains multi-contrast MR images from 23 infants, including T1, T2 images and corresponding labels of Grey Matter (gm) and White Matter (wm).
We randomly select 18 subjects for training and remaining 5 subjects for testing.
661 training images are trained for 800 epochs with batch size 1. 163 images from the 5 subjects are used for testing.

(3)\emph{MRBrain13}: The MRBrain13 dataset \citep{Adri2015MRBrainS} contains multi-contrast MR images from 20 subjects, including T1 and T2-Flair images. We randomly choose 16 subjects for training and the remaining 4 for testing. 704 training images are trained for 1200 epochs with batch size 1. 176 images are used for testing.

(4)\emph{ADNI}: The ADNI dataset \citep{nie2016medical} contains T2 and PD images (proton density images, tissues with a higher concentration or density of protons produce the strongest signals and appear the brightest on the image) from 50 subjects. 40 subjects are randomly selected for training and the remaining 10 for testing.
1795 training images are trained for 400 epochs with batch size 1. 455 images are used for testing.

(5)\emph{RIRE}: The RIRE dataset \citep{West1997Comparison} includes T1 and T2 images collected from 19 subjects.
We randomly choose 16 subjects as for training and the rest for testing.
477 training images are trained for 800 epochs with batch size 1. 156 images are used for testing.

\begin{table*}[!ht]
\renewcommand{\arraystretch}{1.2}
\caption{Comparisons of generation performance evaluated by MAE. Our N2N approach outperforms both Random Forest (RF) based method \citep{Jog2013Magnetic} and Context-Aware GAN (CA-GAN) \citep{nie2016medical} method on most datasets.}
\label{table:results_MAE}
\centering
\resizebox{1.0\textwidth}{!}{
\begin{tabular}{c c c c c c c}
\hline
\multirow{2}{*}{Datasets}& \multirow{2}{*}{Transitions} & \multirow{2}{*}{$RF$}& \multirow{2}{*}{$CA$-$GAN$} & \multicolumn{3}{c}{$N2N$} \\
\cline{5-7}
{} & {} & {} & {}  & $cGAN+L1$  & $cGAN$  & $L1$\\
\hline
\multirow{4}{*}{\textit{BraTs2015}}&T1 $\to$  T2 & \bf{6.025(3.795)} &  11.947(3.768) & 8.292(2.599)& 10.692(3.406)&8.654(3.310)\\
{}&T2 $\to$  T1 & \bf{7.921(5.912)}& 16.587(4.917) & 9.937(5.862)& 15.430(5.828)&10.457(7.016)\\
{}&T1 $\to$  T2-Flair &8.176(6.272)& 13.999(3.060) & \bf{7.934(2.665)}&11.671(3.538) &8.462(3.438)\\
{}&T2 $\to$  T2-Flair&\bf{7.318(4.863)}& 12.658(3.070) & 8.858(2.692)&10.469(4.450)&8.950(3.758)\\
\hline
\multirow{2}{*}{\textit{Iseg2017}}&T1 $\to$  T2 &{3.955(1.936)}& 12.175(2.800) & \bf{3.309(1.274)}&8.028(1.505)&{3.860(1.354)} \\
{}&T2 $\to$  T1 &11.466(9.207)& 17.151(5.181) & \bf{9.586(4.886)}&17.311(4.175)&10.591(5.959)\\
\hline
\textit{MRBrain13} &T1 $\to$  T2-Flair &7.609(3.303)& 13.643(3.117) & \bf{6.064(1.997)}&9.906(3.303)&6.505(2.343) \\
\hline
\multirow{2}{*}{\textit{ADNI}}&PD $\to$  T2 &9.485(3.083)& 16.575(4.538) & {6.757(1.250)}&7.211(1.799) &	\bf{4.898(1.451)}\\
{}&T2 $\to$  PD &5.856(2.560)& 17.648(4.679) & \bf{4.590(1.103)}&5.336(1.534)&5.055(1.914)\\
\hline
\multirow{2}{*}{\textit{RIRE}}&T1 $\to$  T2 &38.047(7.813)& 18.625(5.248) &  \bf{5.250(1.274)}&13.690(3.199)&9.105(1.946)\\
{}&T2 $\to$  T1 &17.022(4.300)& 23.374(5.204) & \bf{9.035(2.146)}&13.964(3.640)&{9.105(1.946)}\\
\hline
\end{tabular}}
\end{table*}

\begin{table*}[!ht]
\renewcommand{\arraystretch}{1.2}
\caption{Comparisons of generation performance evaluated by PSNR. Our N2N approach outperforms both Random Forest (RF) based method \citep{Jog2013Magnetic} and Context-Aware GAN (CA-GAN) \citep{nie2016medical} method on most datasets. }
\label{table:results_PSNR}
\centering
\resizebox{1.0\textwidth}{!}{
\begin{tabular}{c c c c c c c}
\hline
\multirow{2}{*}{Datasets}& \multirow{2}{*}{Transitions} & \multirow{2}{*}{$RF$}& \multirow{2}{*}{$CA$-$GAN$} & \multicolumn{3}{c}{$N2N$} \\
\cline{5-7}
{} & {} & {} & {}  & $cGAN+L1$  & $cGAN$  & $L1$\\
\hline
\multirow{4}{*}{\textit{BraTs2015}}&T1 $\to$  T2  &	\bf{24.717(4.415)}& 19.738(2.489) &22.560(2.020)&20.301(2.079)&22.517(2.311)\\
{}&T2 $\to$  T1 & \bf{23.385(5.391)} & 17.462(2.164) &22.518(3.957)&18.507(2.378) &22.374(4.339)\\
{}&T1 $\to$  T2-Flair &\bf{23.222(5.594)}& 19.157(2.573) 	&22.687(1.939) &19.969(2.111)&22.642(2.530)\\
{}&T2 $\to$  T2-Flair &\bf{23.138(4.172)}& 18.848(1.687) & 21.664(2.211)&20.656(2.628) &21.791(2.621)\\
\hline
\multirow{2}{*}{\textit{Iseg2017}}&T1 $\to$  T2 &28.028(3.386)& 21.992(1.812) &\bf{29.979(1.445)}&22.860(1.524)&{28.874(1.886)} \\
{}&T2 $\to$  T1 &22.342(5.532)& 18.401(2.140) &\bf{23.610(3.339)}&18.121(1.560)&23.325(3.692)\\
\hline
\textit{MRBrain13} &T1 $\to$  T2-Flair &24.780(2.728)& 19.503(1.230) & \bf{26.495(2.506)}&22.616(2.238)&26.299(2.536)\\
\hline
\multirow{2}{*}{\textit{ADNI}}&PD $\to$  T2 &24.006(2.088)& 19.008(2.095) & 26.477(1.609)&26.330(2.081) &\bf{29.089(2.143)}\\
{}&T2 $\to$  PD &29.118(3.409)& 18.715(2.147) &\bf{31.014(1.997)}&29.032(2.012)&30.614(2.483)\\
\hline
\multirow{2}{*}{\textit{RIRE}}&T1 $\to$  T2 &12.862(1.261)& 18.248(3.560) & \bf{28.994(2.450)}&21.038(2.330)&28.951(2.814)\\
{}&T2 $\to$  T1 &19.811(1.918)& 16.029(1.522) &\bf{24.043(1.804)}&20.450(1.969)&24.003(1.699)\\
\hline
\end{tabular}}
\end{table*}

\begin{table*}[!ht]
\renewcommand{\arraystretch}{1.2}
\caption{Comparisons of generation performance evaluated by MI. Our N2N approach outperforms both Random Forest (RF) based method \citep{Jog2013Magnetic} and Context-Aware GAN (CA-GAN) \citep{nie2016medical} method on most datasets. }
\label{table:results_MI}
\centering
\resizebox{1.0\textwidth}{!}{
\begin{tabular}{c c c c c c c}
\hline
\multirow{2}{*}{Datasets}& \multirow{2}{*}{Transitions} & \multirow{2}{*}{$RF$}& \multirow{2}{*}{$CA$-$GAN$} & \multicolumn{3}{c}{$N2N$} \\
\cline{5-7}
{} & {} & {} & {}  & $cGAN+L1$  & $cGAN$  & $L1$\\
\hline
\multirow{4}{*}{\textit{BraTs2015}}&T1 $\to$  T2 & 0.617(0.239) &  0.787(0.075) & 0.862(0.080)& 0.788(0.078)&\bf{0.901(0.085)}\\
{}&T2 $\to$  T1 & 0.589(0.217)& 0.661(0.074) & 0.777(0.077)& 0.673(0.061)&\bf{0.818(0.075)}\\
{}&T1 $\to$  T2-Flair &0.609(0.225)& 0.722(0.059) & 0.833(0.068)&0.749(0.057) &\bf{0.879(0.078)}\\
{}&T2 $\to$  T2-Flair&0.610(0.230)& 0.756(0.062) & 0.848(0.063)&0.817(0.065)&\bf{0.928(0.069)}\\
\hline
\multirow{2}{*}{\textit{Iseg2017}}&T1 $\to$  T2 &0.803(0.306)& 0.804(0.172) & 0.931(0.179)&0.782(0.149)&\bf{0.993(0.183)} \\
{}&T2 $\to$  T1 &0.788(0.299)& 0.789(0.201) & 0.868(0.214)&0.777(0.166)&\bf{0.880(0.198)}\\
\hline
\textit{MRBrain13} &T1 $\to$  T2-Flair &1.123(0.175)& 0.805(0.252) & 1.066(0.121)&1.009(0.082)&\bf{1.185(0.093)} \\
\hline
\multirow{2}{*}{\textit{ADNI}}&PD $\to$  T2 &1.452(0.117)& 0.674(0.199) & 1.266(0.124)&1.184(0.113) &	\bf{1.484(0.140)}\\
{}&T2 $\to$  PD &1.515(0.154)& 0.659(0.196) & 1.381(0.172)&1.282(0.120)&\bf{1.536(0.150)}\\
\hline
\multirow{2}{*}{\textit{RIRE}}&T1 $\to$  T2 &0.694(0.192)& \bf{0.724(0.113) }&  0.636(0.191)&0.513(0.141)&0.698(0.194)\\
{}&T2 $\to$  T1 &0.944(0.130)& 0.650(0.226) & 0.916(0.137)&0.737(0.101)&\bf{0.969(0.142)}\\
\hline
\end{tabular}}
\end{table*}

\begin{table*}[!ht]
\renewcommand{\arraystretch}{1.2}
\caption{Comparisons of generation performance evaluated by SSIM. Our N2N approach outperforms both Random Forest (RF) based method \citep{Jog2013Magnetic} and Context-Aware GAN (CA-GAN) \citep{nie2016medical} method on most datasets. }
\label{table:results_SSIM}
\centering
\resizebox{1.0\textwidth}{!}{
\begin{tabular}{c c c c c c c}
\hline
\multirow{2}{*}{Datasets}& \multirow{2}{*}{Transitions} & \multirow{2}{*}{$RF$}& \multirow{2}{*}{$CA$-$GAN$} & \multicolumn{3}{c}{$N2N$} \\
\cline{5-7}
{} & {} & {} & {}  & $cGAN+L1$  & $cGAN$  & $L1$\\
\hline
\multirow{4}{*}{\textit{BraTs2015}}&T1 $\to$  T2 &  \bf{0.910(0.050) } &  0.826(0.022) &  0.866(0.029) &  0.575(0.046) & 0.880(0.029) \\
{}&T2 $\to$  T1 &  0.893(0.060)  &  0.723(0.027) &  0.854(0.054) &  0.723(0.027) & \bf{0.896(0.037) }\\
{}&T1 $\to$  T2-Flair &  \bf{0.873(0.072) } &  0.756(0.025) &  0.837(0.025) &  0.797(0.027) & 0.857(0.028) \\
{}&T2 $\to$  T2-Flair&  \bf{0.875(0.066) } &  0.749(0.016) &  0.836(0.022) &  0.823(0.031) & 0.860(0.026) \\
\hline
\multirow{2}{*}{\textit{Iseg2017}}&T1 $\to$  T2 &  0.902(0.054)  &  0.690(0.149) &  0.887(0.034) &  0.748(0.102) & \bf{0.913(0.030) } \\
{}&T2 $\to$  T1 &  \bf{0.808(0.112 } &  0.662(0.144) &  0.745(0.137) &  0.620(0.102) & 0.754(0.135) \\
\hline
\textit{MRBrain13} &T1 $\to$  T2-Flair &  0.863(0.058)  &  0.782(0.054) &  0.823(0.074) &  0.785(0.066) & \bf{0.881(0.058) } \\
\hline
\multirow{2}{*}{\textit{ADNI}}&PD $\to$  T2 &  0.819(0.093)  &  0.728(0.045) &  0.812(0.033) &  0.779(0.048) & \bf{0.891(0.042) }\\
{}&T2 $\to$  PD &  {0.880(0.076) } &  0.713(0.053) &  0.856(0.047) &  0.820(0.031) &\bf{0.881(0.066) }\\
\hline
\multirow{2}{*}{\textit{RIRE}}&T1 $\to$  T2 &  0.501(0.0820)  &  0.749(0.087) &  0.736(0.047) &  0.506(0.027) & \bf{0.760(0.045) }\\
{}&T2 $\to$  T1 &  0.622(0.074)  &  0.728(0.112) &  0.692(0.058) &  0.538(0.058) & \bf{0.741(0.048) }\\
\hline
\end{tabular}}
\end{table*}

\begin{figure*}[!ht]
\centering
\includegraphics[width=\textwidth]{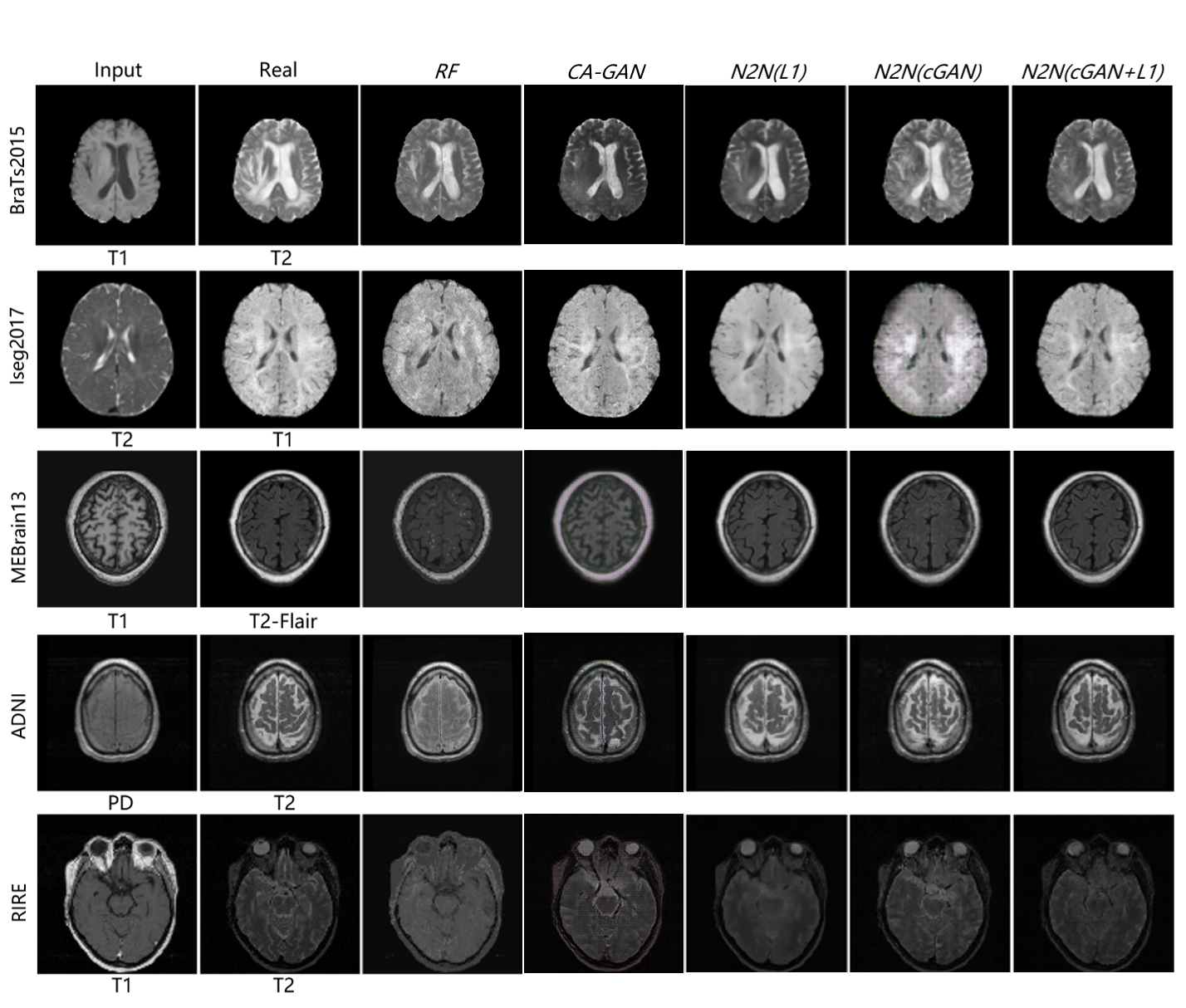}
\caption{Samples of cross-modality generation results on five publicly available datasets including \textit{BraTs2015} \citep{Menze2015The}, \textit{Iseg2017} \citep{wang2015links}, \textit{MRBrain13} \citep{Adri2015MRBrainS}, \textit{ADNI} \citep{nie2016medical}, and \textit{RIRE} \citep{West1997Comparison}. Results are selected from top performing examples (relatively low MAE, high PSNR, high MI, and high PSNR collectively ) with four approaches. The right five columns show results of the random-forests-based method (RF) \citep{Jog2013Magnetic}, the Context-Aware GAN (CA-GAN) \citep{nie2016medical} and N2N framework with different loss functions ($L1, cGAN, cGAN+L1$).}
\label{fig:results_generation}
\end{figure*}

\textbf{Results.}
Generation performance with different methods on the five datasets are summarized in Table \ref{table:results_MAE}, Table \ref{table:results_PSNR}, Table \ref{table:results_MI} and Table \ref{table:results_SSIM}.
It quantitatively shows how using N2N translation network allows us to achieve better generation results than the regression-based method using RF \citep{Jog2013Magnetic} and the latest proposed Context-Aware GAN method from \citep{nie2016medical} on most datasets evaluated by MAE, PSNR, MI, and SSIM. However, there are also some cases where the RF method surpasses our N2N translation network on the \textit{BraTs2015} dataset (images with tumors). It is explicable since the RF method incorporates additional context features, taking full advantages of structural information and thus leading to comparable generation results on images with tumors.

Note that different losses induce different quality of generated images. In most cases, our N2N network with $cGAN+L1$ achieves the best results on MAE and PSNR; $L1$ loss term contributes to superior performance on MI over other methods. MI focuses more attention on the matching of pixel-wise intensities and ignores structural information in the images. Meanwhile, the L1 loss term ensures pixel-wise information rather than the properties of human visual perception \citep{Larsen2015Autoencoding}. Thus, it is reasonable that using L1 term contributes to superior results on MI.
\begin{table}[!ht]
\renewcommand{\arraystretch}{1.2}
\caption{Segmentation results of N2N translated images on \textit{BraTs2015} evaluated by FCN-score. The gap between translated images and the real images can evaluate the generation performance of our method. Note that ``all'' represents mean accuracy of all pixels (the meanings of ``all'' are the same in the following tables). We achieve close segmentation results between translated-modality images and target-modality images.}
\label{table:results_brats_fcnscore}
\centering
\scriptsize
\begin{tabular}{c c c c}
\hline
\multirow{2}{*}{Method} & \multicolumn{2}{c }{Accuracy} & {Dice} \\
\cline{2-4}
{} & all  & tumor  & tumor  \\
\hline
T1 $\to$  T2 & 0.955 &0.716 &0.757   \\
T2 (real)      & 0.965 &0.689 &0.724  \\
\hline
T2 $\to$  T1 &0.958 &0.663 &0.762  \\
T1 (real)      &0.972 &0.750 &0.787   \\
\hline
T1 $\to$  T2-Flair &0.945 &0.729 &0.767  \\
T2 $\to$  T2-Flair &0.966 &0.816 &0.830  \\
T2-Flair (real)      &0.986 &0.876 &0.899   \\
\hline
\end{tabular}
\end{table}

\begin{table}[!htp]
\renewcommand{\arraystretch}{1.2}
\caption{Segmentation results of N2N translated images on \textit{Iseg2017} evaluated by FCN-score. Note that ``gm'' and ``wm'' indicate gray matter and white matter respectively. The minor gap between translated-modality images and the target-modality images shows decent generation performance of our framework. }
\label{table:results_iseg_fcnscore}
\centering
\scriptsize
\begin{tabular}{c c c c c c}
\hline
\multirow{2}{*}{Method} & \multicolumn{3}{c }{Accuracy} & \multicolumn{2}{c}{Dice} \\
\cline{2-6}
{} & all & gm & wm  & gm & wm  \\
\hline
T1 $\to$ T2 & 0.892 & 0.827 & 0.506 & 0.777 & 0.573  \\
T2 (real)      & 0.920 & 0.829 & 0.610 & 0.794 & 0.646 \\
\hline
T2 $\to$ T1 & 0.882 & 0.722 & 0.513 & 0.743  & 0.569 \\
T1 (real)      & 0.938 & 0.811 & 0.663 & 0.797 & 0.665 \\
\hline
\end{tabular}
\end{table}

Fig.\ref{fig:results_generation} shows the qualitative results of cross-modality image generation using different approaches on five datasets. We have reasonable but blurry results using N2N network with $L1$ alone. The N2N network with $cGAN$ alone leads to improvements in visual performance but causes some artifacts in cross-modality MR image generation. Using $cGAN+L1$ terms achieves decent results and reduces artifacts. In contrast, the RF method and Context-Aware GAN lead to rough and fuzzy results compared with N2N networks.

We also quantify the generation results using FCN-score on \textit{BraTs2015} and \textit{Iseg2017} in Table \ref{table:results_brats_fcnscore} and Table \ref{table:results_iseg_fcnscore}. Our approach ($cGAN+L1$) is effective in generating realistic cross-modality MR images towards the real images. The cGAN-based objectives lead to high scores close to the real images.

To validate the perceptual realism of our generated images, two more experiments are conducted. One is conducted by three radiologists. The other is done by five well-trained medical students. For the first experiment, we randomly select 1100 pairs of images, each of which consists of an image generated by our framework and its corresponding real image. On each trial, three radiologists are respectively asked to select which one is fake in the image pair. The first 100 trials are practice after which they are given feedback. The following 1000 trials are the main experiment where no feedback are given. The average performance of the three radiologists quantitatively evaluates the perceptual realism of our approach. For the second experiment, the experimental setting is perfectly identical.
Results indicate that our generated images fooled radiologists on 25\% trials and fooled students on 27.6\% trials.

\subsection{Cross-Modality Registration}
\textbf{Evaluation metric.} We use the two evaluation metrics for cross-modality registration, namely Dice and Distance Between Corresponding Landmarks (Dist).

(1)\emph{Dice}: The first metric is introduced to measure the overlap of ground truth segmentation labels and registered segmentation labels. It is defined as $(2 |H\cap G|)/(|H|+|G|)$ where $G$ is the ground truth segmentation label of the fixed image and $H$ is the registered segmentation label of the fixed image. Since image registration involves identification of a transformation to fit a fixed image to a moving image. The success of the registration process is vital for correct interpretation of many medical image-processing applications, including multi-atlas segmentation. A higher Dice, which measures the overlap of propagated segmentation labels through deformation and the ground truth labels, indicates a more accurate registration.

(2)\emph{Distance Between Corresponding Landmarks (Dist)}: The second metric is adopted to measure the capacity of algorithms to register the brain structures. The registration error on a pair of images is defined as the average Euclidean distance between a landmark in the warped image and its corresponding landmark in the fixed image. To compute the Euclidean distance, all 2D-slices after registration are stacked into 3D images.

\textbf{Dataset.} We preprocess the original MRI data from \textit{Iseg2017} and \textit{MRBrain13} datasets with the following steps to make it applicable to our proposed framework. (1) We first shear the 3D image into a smaller cube, each side of which circumscribes the brain. (2) The brain cube is then resized to a size of $128\times128\times128$ voxels. (3) The last step is to slice the brain cubes from all the subjects into 2D data along the z-axis ($128\times128$, 128 slices).

After preprocessing, the brain slices with the same depth value from different subjects are spatially aligned. During the training phase, a pair of brain slices from two different subjects with the same depth value is treated as a pair moving and fixed images. In order to conduct five-fold cross-validation for our experiments, the value of $n$ (numbers of atlases) is selected differently in each dataset. For \textit{Iseg2017} dataset, we choose 8 subjects in the moving images space and another 2 subjects in the fixed images space ($n=8$). For \textit{MRBrain13} dataset, 4 subjects are selected for the moving images space while one subject in the fixed images space ($n=4$)

\begin{figure}[!htp]
\centering
\includegraphics[width=80mm]{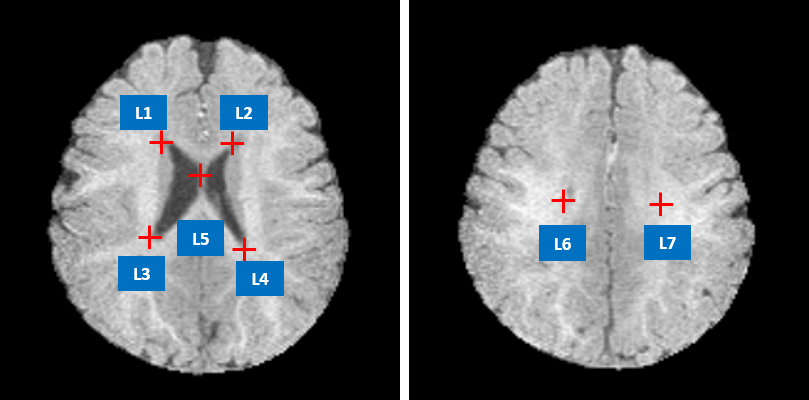}
\caption{Illustration of the seven landmarks selected for cross-modality registration. L1: right lateral ventricle superior, L2: left lateral ventricle superior, L3: right lateral ventricle inferior, L4: left lateral ventricle inferior. L5: middle of the lateral ventricle, L6: right lateral ventricle posterior, L7: left lateral ventricle posterior.}
\label{fig:landmark}
\end{figure}

\textit{Iseg2017} and \textit{MRBrain13} datasets provide ground truth segmentation labels. Seven well-defined anatomic landmarks (see Fig.\ref{fig:landmark}) that are distributed in the lateral ventricle are manually annotated by three doctors. We consider the average coordinates from three doctors as the ground truth positions of the landmarks.

\begin{table*}[!ht]
\renewcommand{\arraystretch}{1.2}
\caption{Registration results evaluated by Dist and Dice on \textit{Iseg2017} and \textit{MRBrain13}.   }
\label{table:results_registration}
\centering
\scriptsize
\resizebox{1.0\textwidth}{!}{
\begin{tabular}{c c c c c c c}
\hline
\multirow{2}{*}{Datasets}&\multirow{2}{*}{Modalities } & \multirow{2}{*}{Structures} & \multicolumn{2}{c}{Dice} & \multicolumn{2}{c}{Dist}\\
\cline{4-7}
{}&{} & {}  & {ANTs}& {Elastix} & {ANTs}& {Elastix} \\
\hline
\multirow{8}{*}{\textit{Iseg2017}} & \multirow{2}{*}{T2} &wm  &0.508$\pm$0.008 & 0.475$\pm$0.006 &\multirow{2}{*}{2.105$\pm$0.006} &\multirow{2}{*}{ 2.836$\pm$0.014}  \\

{}&{}&{gm}&0.635$\pm$0.015 &0.591$\pm$0.014 &{} &{} \\

{} & \multirow{2}{*}{$\widehat {T}1$} &wm  &0.503$\pm$0.004 & 0.469$\pm$0.005 &\multirow{2}{*}{1.884$\pm$0.011} &\multirow{2}{*}{2.792$\pm$0.008}   \\

{}&{}&{gm}&0.622$\pm$0.014&0.580$\pm$0.012 &{} &{} \\

{} & \multirow{2}{*}{T2$+\widehat{T}1$} &wm  &\bf{0.530$\pm$0.009} &\bf{0.519$\pm$0.007} &\multirow{2}{*}{\bf{1.062$\pm$0.017}} &\multirow{2}{*}{\bf{2.447$\pm$0.009}}  \\

{}&{}&{gm}&\bf{0.657$\pm$0.016}&\bf{0.648$\pm$0.015}&{} &{} \\
{} & \multirow{2}{*}{T1} &wm  &0.529$\pm$0.008 & 0.500$\pm$0.014 &\multirow{2}{*}{1.136$\pm$0.009} &\multirow{2}{*}{2.469$\pm$0.012}  \\

{}&{}&{gm}& 0.650$\pm$0.016 &0.607$\pm$0.018 &{} &{} \\

{} & \multirow{2}{*}{$\widehat {T}2$} &wm  &0.495$\pm$0.007 & 0.457$\pm$0.005 &\multirow{2}{*}{2.376$\pm$0.013} &\multirow{2}{*}{3.292$\pm$0.011}   \\

{}&{}&{gm}&0.617$\pm$0.017&0.573$\pm$0.012 &{} &{} \\
{} & \multirow{2}{*}{T1+$\widehat{T}2$} &wm  &\bf{0.538$\pm$0.009} &\bf{0.527$\pm$0.006} &\multirow{2}{*}{\bf{1.097$\pm$0.008}} &\multirow{2}{*}{\bf{2.116$\pm$0.009}}  \\

{}&{}&{gm}&\bf{0.664$\pm$0.017}&\bf{0.650$\pm$0.017} &{} &{} \\

{} & \multirow{2}{*}{T1+T2} &wm  &0.540$\pm$0.009 &0.528$\pm$0.006 &\multirow{2}{*}{1.013$\pm$0.007} &\multirow{2}{*}{2.109$\pm$0.008}  \\

{}&{}&{gm}&0.666$\pm$0.017&0.651$\pm$0.017&{} &{} \\
\hline
\multirow{8}{*}{\textit{MRBrain13}} & \multirow{2}{*}{T2-Flair} &wm  &0.431$\pm$0.025  & 0.412$\pm$0.010   &\multirow{2}{*}{3.417$\pm$0.031 } &\multirow{2}{*}{3.642$\pm$0.023 }    \\

{}&{}&{gm}&0.494$\pm$0.026 &0.463$\pm$0.023 &{} &{} \\

{} & \multirow{2}{*}{$\widehat {T}1$} &wm  &0.468$\pm$0.032 &0.508$\pm$0.012   &\multirow{2}{*}{3.159$\pm$0.016 } &\multirow{2}{*}{3.216$\pm$0.014 }  \\

{}&{}&{gm}&0.508$\pm$0.024 &0.487$\pm$0.018 &{} &{} \\

{} & \multirow{2}{*}{T2-Flair+$\widehat{T}1$} &wm  &\bf{0.473$\pm$0.027}  & \bf{0.492$\pm$0.012}  &\multirow{2}{*}{\bf{2.216$\pm$0.011} } &\multirow{2}{*}{\bf{2.659$\pm$0.021} }  \\

{}&{}&{gm}&\bf{0.530$\pm$0.027} &\bf{0.532$\pm$0.029} &{} &{} \\

{} & \multirow{2}{*}{T1} &wm  &0.484$\pm$0.038  & 0.534$\pm$0.005  &\multirow{2}{*}{2.524$\pm$0.022 } &\multirow{2}{*}{ 2.961$\pm$0.019} \\

{}&{}&{gm}&0.517$\pm$0.025 &0.510$\pm$0.018 &{} &{} \\

{} & \multirow{2}{*}{$\widehat {T}2$-$Flair$} &wm  &0.431$\pm$0.022 &0.410$\pm$0.012   &\multirow{2}{*}{3.568$\pm$0.039 } &\multirow{2}{*}{3.726$\pm$0.024 }  \\

{}&{}&{gm}&0.497$\pm$0.018 &0.458$\pm$0.018 &{} &{} \\

{} & \multirow{2}{*}{T1+$\widehat{T}2$-$Flair$} &wm  &\bf{0.486$\pm$0.033}  & \bf{0.505$\pm$0.011}  &\multirow{2}{*}{\bf{2.113$\pm$0.014} } &\multirow{2}{*}{\bf{2.556$\pm$0.020} }  \\

{}&{}&{gm}&\bf{0.534$\pm$0.025} &\bf{0.540$\pm$0.029} &{} &{} \\

{} & \multirow{2}{*}{T2-Flair+T1} &wm  &0.486$\pm$0.033 & 0.503$\pm$0.013  &\multirow{2}{*}{2.098$\pm$0.013 } &\multirow{2}{*}{2.508$\pm$0.019 }  \\

{}&{}&{gm}&0.534$\pm$0.027 &0.539$\pm$0.029 &{} &{} \\
\hline
\end{tabular}}
\end{table*}

\textbf{Results.}
Our experiments not only include registration with real data, but also with translated images ($\widehat {T}1$ and $\widehat {T}2$ images for \textit{Iseg2017} dataset, $\widehat {T}1$ and $\widehat{T}2$-$Flair$ images for \textit{MRBrain13} dataset). The deformations generated in each set of experiments are combined in a weighted fusion process, yielding the final registration deformation. In order to compute the Euclidean distance of those corresponding landmarks between warped images and fixed images, all 2D-slices are then stacked into 3D images. Besides, we also employ the fused deformation to segmentation labels of moving images, obtaining registered segmentation results of fixed images.

Table \ref{table:results_registration} summarizes the registration results both in terms of Dist and Dice. We introduce the cross-modality information from our $\widehat {T}1$ images into T2 images and T2-Flair images, of which the performance are denoted as ``T2+$\widehat {T}1$'' and ``T2-Flair+$\widehat {T}1$''. Likewise, ``T1+$\widehat {T}2$'' and ``T1+$\widehat{T}2$-$Flair$'' indicate performance of registrations with cross-modality information from our $\widehat{T}2$-$Flair$ images added into T1 images. We also show the upper bounds of registrations with translated images, which are denoted as ``T1+T2'' and ``T2-Flair+T1''. The weights for the combination are determined through five-fold cross-validation. The optimal weights of 0.92 and 0.69 are selected for $\hat{T}1$ images in terms of white matter and gray matter on \textit{Iseg2017} and 0.99 and 0.82 are selected on \textit{MRBrain13}.

\begin{figure}[!htp]
\centering
\includegraphics[width=\textwidth]{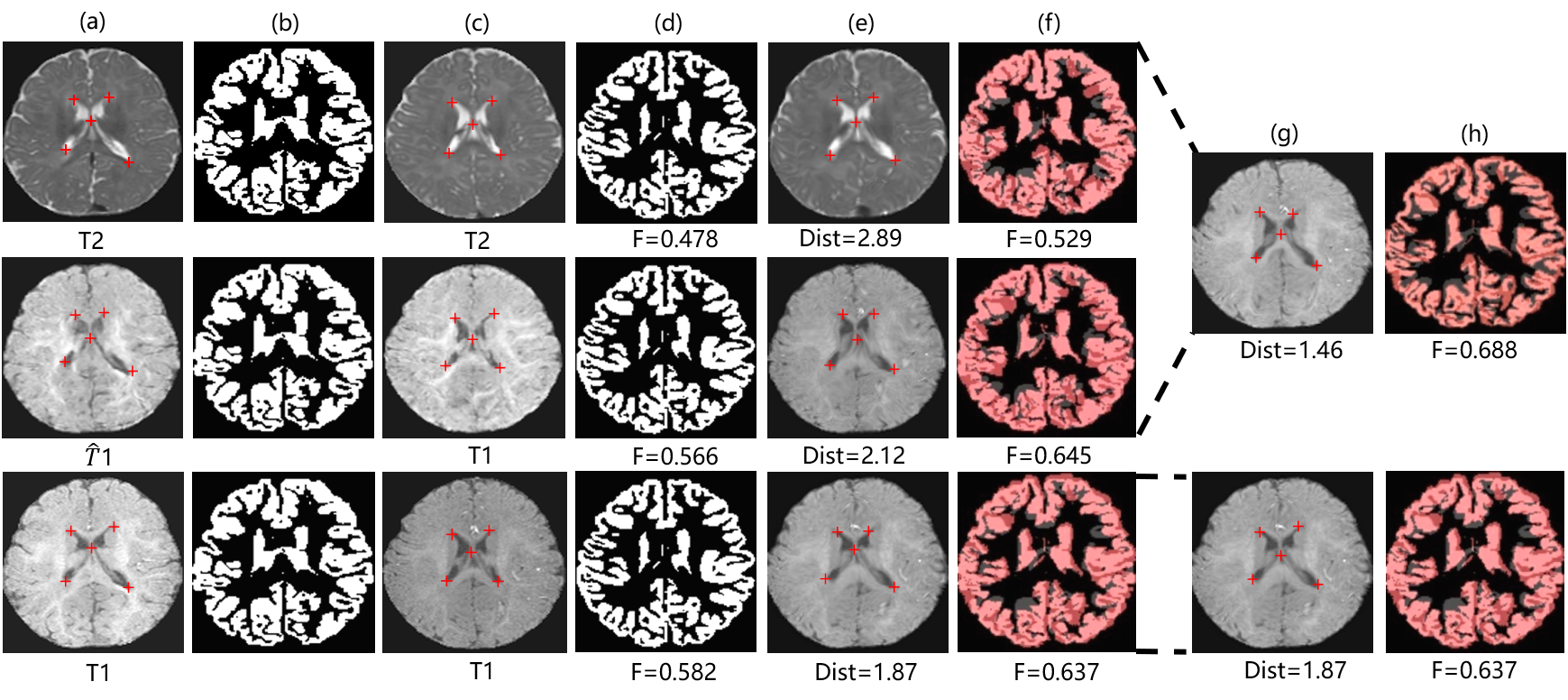}
\caption{Samples of registration results of our method: (a) Fixed image, (b) Ground truth segmentation label of fixed image, (c) Moving image, (d) Ground truth segmentation label of moving image, (e) Warped image (moving image warped by the best traditional registration algorithm (ANTs)), (f) Warped ground truth segmentation label of moving image, (g) Fused image, (h) Segmentation prediction of fused image. The Blue, dark blue, grey areas in (f) denote true regions, false regions, and missing regions respectively. The red crosses denote landmarks in the fixed and moving images. }
\label{fig:registration-visual}
\end{figure}

After the weighted fusion process, we find that registrations with translated images show better performance than those with real data by achieving higher Dice, e.g. 0.657$\pm$0.016 (T2+$\widehat {T}1$) vs. 0.635$\pm$0.015 (T2) and 0.534$\pm$0.025 (T1+$\widehat{T}2$-$Flair$) vs. 0.517$\pm$0.025 (T1). We also observe that the Dist is greatly shortening (e.g. 2.216$\pm$0.011 (T2-Flair+$\widehat {T}1$) vs. 3.417$\pm$0.031(T2-Flair)) compared to registrations without adding cross-modality information. In many cases, our method even advances the upper bound both in Dist and Dice. These results are reasonable because our translated images are realistic enough, as well as the real data itself with high contrast for brain structure leads to lower registration errors. Fig.\ref{fig:registration-visual} visualizes samples of the registration results of our methods. More details can be found there.

\begin{table*}[!ht]
\renewcommand{\arraystretch}{1.2}
\caption{Results of our additional registration experiments evaluated by Dist and Dice on \textit{Iseg2017} and \textit{MRBrain13} realized by ANTS.   }
\label{table:results_registration2}
\centering
\scriptsize
\resizebox{0.78\textwidth}{!}{
\begin{tabular}{c c c c c }
\hline
Datasets&Modalities  & Structures & Dice &Dist\\
\hline
\multirow{14}{*}{\textit{Iseg2017}} & \multirow{2}{*}{T2} &wm  &0.823$\pm$0.283 & \multirow{2}{*}{0.475$\pm$0.006} \\

{}&{}&{gm}&0.859$\pm$0.227 &{}  \\

{} & \multirow{2}{*}{$\widehat {T}1$} &wm  &0.882$\pm$0.254 &\multirow{2}{*}{0.183$\pm$0.167}   \\

{}&{}&{gm}&0.910$\pm$0.195&{}  \\

{} & \multirow{2}{*}{T2$+\widehat{T}1$} &wm  &\bf{0.883$\pm$0.252} &\multirow{2}{*}{\bf{0.190$\pm$0.171}}  \\

{}&{}&{gm}&\bf{0.657$\pm$0.911}&{}\\
{} & \multirow{2}{*}{T1} &wm  &0.868$\pm$0.263 & \multirow{2}{*}{0.179$\pm$0.085}  \\

{}&{}&{gm}& 0.898$\pm$0.206 &{} \\

{} & \multirow{2}{*}{$\widehat {T}2$} &wm  &0.807$\pm$0.295 & \multirow{2}{*}{0.218$\pm$0.416}  \\

{}&{}&{gm}&0.846$\pm$0.203&{} \\
{} & \multirow{2}{*}{T1+$\widehat{T}2$} &wm  &\bf{0.868$\pm$0.259} &\multirow{2}{*}{\bf{0.186$\pm$0.095}} \\

{}&{}&{gm}&\bf{0.898$\pm$0.198}&{}  \\

{} & \multirow{2}{*}{T1+T2} &wm  &0.868$\pm$0.256 &\multirow{2}{*}{0.184$\pm$0.089}  \\

{}&{}&{gm}&0.898$\pm$0.201&{}\\
\hline
\multirow{14}{*}{\textit{MRBrain13}} & \multirow{2}{*}{T2-Flair} &wm  &0.976$\pm$0.116  & \multirow{2}{*}{0.182$\pm$0.083}   \\

{}&{}&{gm}&0.976$\pm$0.132 &{}  \\

{} & \multirow{2}{*}{$\widehat {T}1$} &wm  &0.966$\pm$0.157 &\multirow{2}{*}{0.181$\pm$0.086}   \\

{}&{}&{gm}&0.968$\pm$0.162 &{} \\

{} & \multirow{2}{*}{T2-Flair+$\widehat{T}1$} &wm  &\bf{0.971$\pm$0.105}  & \multirow{2}{*}{\bf{0.180$\pm$0.086}}  \\

{}&{}&{gm}&\bf{0.974$\pm$0.095} &{} \\

{} & \multirow{2}{*}{T1} &wm  &0.976$\pm$0.127  & \multirow{2}{*}{0.179$\pm$0.085}\\

{}&{}&{gm}&0.981$\pm$0.123 &{} \\

{} & \multirow{2}{*}{$\widehat {T}2$-$Flair$} &wm  &0.985$\pm$0.079 &\multirow{2}{*}{0.180$\pm$0.085}   \\

{}&{}&{gm}&0.983$\pm$0.109 &{} \\

{} & \multirow{2}{*}{T1+$\widehat{T}2$-$Flair$} &wm  &\bf{0.985$\pm$0.051}  & \multirow{2}{*}{\bf{0.179$\pm$0.085} }  \\

{}&{}&{gm}&\bf{0.985$\pm$0.062} &{} \\

{} & \multirow{2}{*}{T2-Flair+T1} &wm  &0.978$\pm$0.081 & \multirow{2}{*}{0.178$\pm$0.085}  \\

{}&{}&{gm}&0.982$\pm$0.076 &{} \\
\hline
\end{tabular}}
\end{table*}
 To demonstrate the effectiveness of our cross-modality registration approach with translated images, we propose an additional experiment by employing a known transformation to the moving images to generate transformed images that can be used as our ``fixed''. This allows us to directly estimate the benefit of adding translated modalities to the registration process when finding the known transformation during the registration step. Take T1 and T2 images as one example. The T1 and T2 images from the moving images space are first rotated a certain degree. Here we rotate them by 30 degrees. The $\hat{T}1$ images generated from our framework are also rotated 30 degrees. All these rotated images are used as our ``fixed''. T2 (moving) images are registered to rotated T2 (fixed) images and T1 (moving) images are registered to rotated $\hat{T}1$ (fixed) images. The following fusion processes are the same as our stated method. Table \ref{table:results_registration2} shows the results of our additional experiments.

\subsection{Cross-Modality Segmentation}
\textbf{Evaluation metric.} We report segmentation results on Dice (higher is better).

\textbf{Dataset.} The original training set is divided into $Part A$ and $Part B$ at the ratio of 1:1 based on the subjects. The original test set maintains the same (denoted as $Part C$). $Part A$ is used to train the generator. $Part B$ is then used to infer the translated modality. $Part B$ is then used to train the segmentation model, which is tested on $Part C$.

(1)\emph{Brats2015}: The original \textit{Brats2015} dataset contains 1924 images ($Part A$: 945, $Part B$: 979) for training and 451 images ($Part C$) for testing. After preprocessing, 979 images are trained for 400 epochs and 451 images are used for testing.

(2)\emph{Iseg2017}: The original \textit{Iseg2017} dataset contains 661 images ($Part A$: 328, $Part B$:333) for training and 163 images ($Part C$) for testing. After preprocessing, 333 images are trained for 800 epochs and 163 images remain for testing.

\textbf{Results. }Our experiments focus on two types of MRI brain segmentation: tumor segmentation and brain structure segmentation. Among all MRI modalities, some modalities are conducive to locating tumors (e.g. T2 and T2-Flair) and some are utilized for observing brain structures (e.g. T1) like white matters and gray matters. To this point, we choose to add cross-modality information from T2 and T2-Flair images into T1 images for tumor segmentation and add cross-modality information from T1 images into T2 images for brain structure segmentation. Experiments of tumor segmentation are conducted on \textit{Brats2015} and experiments of brain structure segmentation are conducted on \textit{Iseg2017}.
\begin{table*}[!ht]
\renewcommand{\arraystretch}{1.2}
\caption{Tumor segmentation results of TMS on \textit{Brats2015}. ``T1+$\widehat {T}2$'' and ``T1+$\widehat {T}2\text{-}Flair$'' indicate our approach (TMS) where inputs are both T1 and $\widehat {T}2$ images or T1 and $\widehat {T}2\text{-}Flair$ images. ``T1'' indicates the traditional FCN method where inputs are only T1 images. ``T1+T2'' and ``T1+T2-Flair'' indicate the upper bound. $\Delta$ indicates the increment between TMS and the the traditional FCN method.  }
\label{table:results_brats_seg}
\centering
\scriptsize
\begin{tabular}{c c c}
\hline
{} & {Dice(tumor)} & $\Delta$  \\
\hline
T1 &0.760 & - \\
\bf{T1}$\mathbf{+\widehat {T}2}$ & \bf{0.808}& \bf{6.32\%} \\
T1+T2 & 0.857& - \\
\bf{T1}$\mathbf{+\widehat {T}2\text{-}Flair}$ &\bf{0.819} & \bf{7.89\%} \\
T1+T2-Flair & 0.892& - \\
\hline
\end{tabular}
\end{table*}

\begin{table*}[!ht]
\renewcommand{\arraystretch}{1.2}
\caption{Brain structure segmentation results of TMS on \textit{Iseg2017}. ``T2+$\widehat {T}1$'' indicates our method (TMS) where inputs are both T2 and $\widehat {T}1$ images. ``T2'' indicates the traditional FCN method where inputs are only T2 images. ``T2+T1'' indicates the upper bound. }
\label{table:results_iseg2017_seg}
\centering
\scriptsize
\begin{tabular}{c c c c c}
\hline
{} & Dice(wm)  & $\Delta$ & Dice(gm) & $\Delta$  \\
\hline
T2 &0.649 & -& 0.767 & - \\
\bf{T2}$\mathbf{+\widehat{T}1}$ & \bf{0.669}& \bf{3.08\%} &\bf{0.783} &\bf{2.09\%} \\
T2+T1 & 0.691& - &0.797 & -\\
\hline
\end{tabular}
\end{table*}
\begin{figure*}[!htp]
\centering
\includegraphics[width=\textwidth]{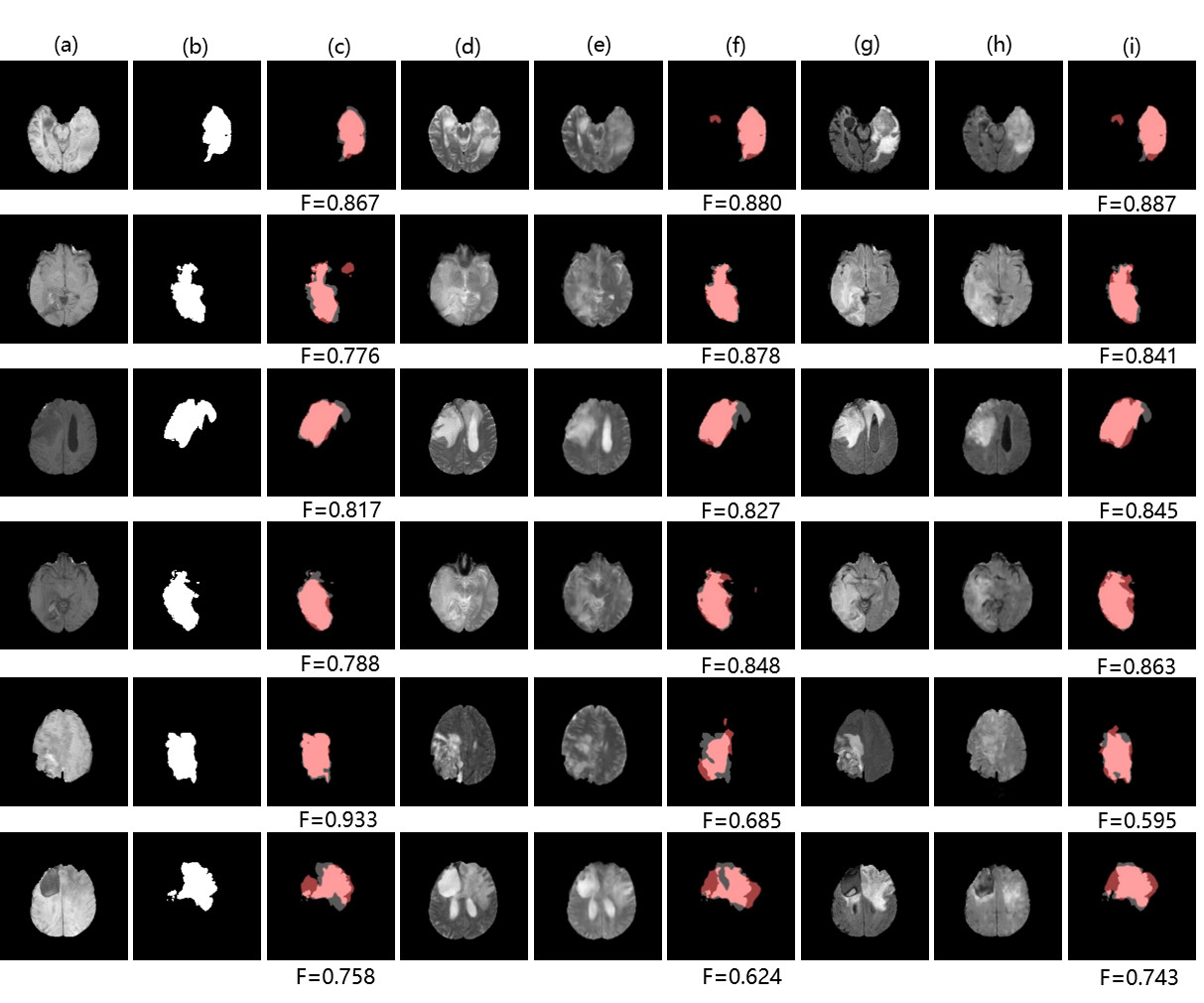}
\caption{Samples of tumor segmentation results on \textit{BraTs2015}: (a), (d), (e), (g), (h) denote T1 image, T2 image, $\widehat {T}2$ image, T2-Flair image, $\widehat {T}2\text{-}Flair$ image. (b) denotes ground truth segmentation label of T1 image. (c), (f), (i) denote tumor segmentation results of T1 image using the FCN method, TMS (adding cross-modality information from $\widehat {T}2$ image), TMS (adding cross-modality information from $\widehat {T}2\text{-}Flair$ image). Note that we have four decent samples in the first four rows and two abortive cases in the last two rows. Pink: true regions. Grey: missing regions. Dark red: false regions.}
\label{fig:results_segmentation}
\end{figure*}

\begin{figure*}[!htp]
\centering
\includegraphics[width=\textwidth]{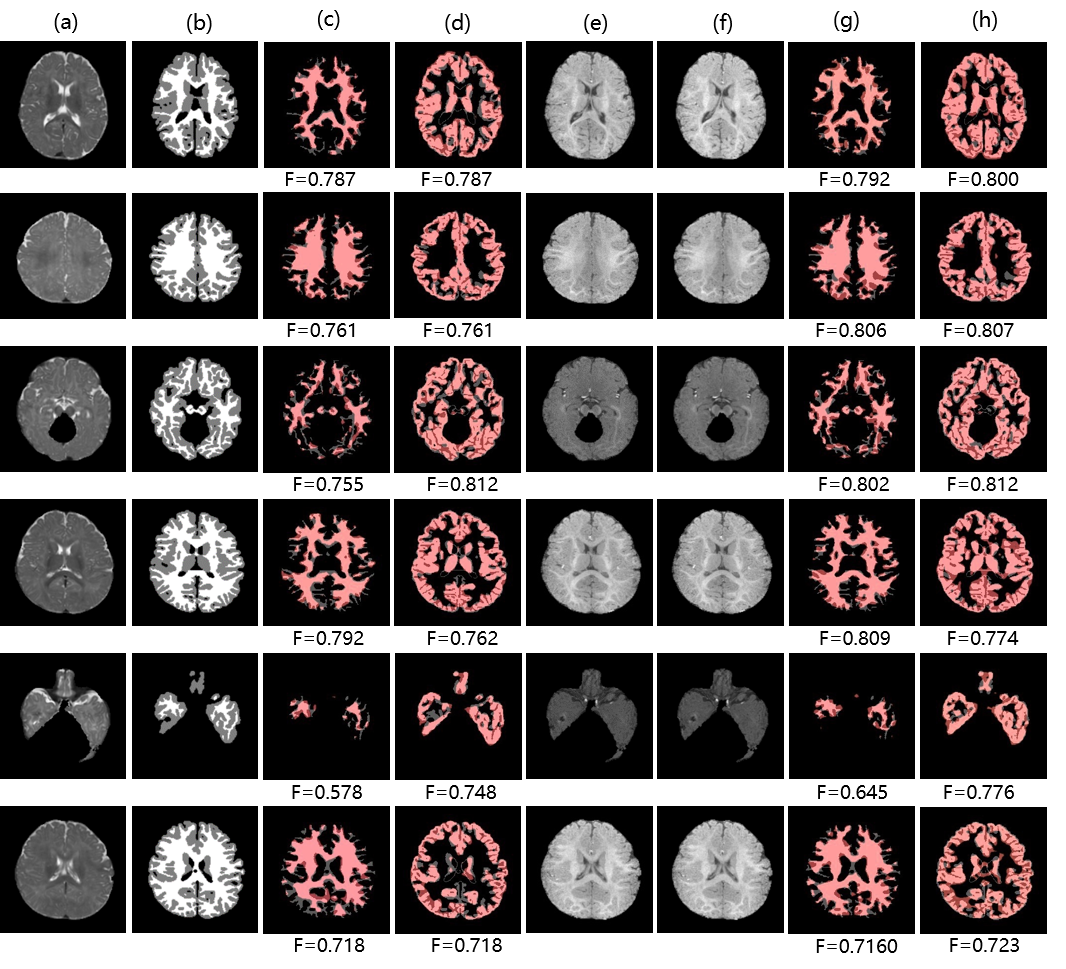}
\caption{Samples of brain structure segmentation results on \textit{Iseg2017}: (a), (e), (f) denote T2 image, T1 image, $\widehat {T}1$ image. (b) denotes ground truth segmentation label of T2 image. (c), (d) denote white matter and gray matter segmentation results of T2 image using the FCN method respectively. (g), (h) denote white matter and gray matter segmentation results of T2 image using TMS (adding cross-modality information from $\widehat {T}1$ image) respectively. Note that we have four decent samples in the first four rows and two abortive cases in the last two rows. Pink: true regions. Grey:  missing regions. Dark red: false regions.}
\label{fig:results_segmentation_seg}
\end{figure*}

As shown in Tables \ref{table:results_brats_seg}, cross-modality information from $\widehat {T}2\text{-}Flair$ and $\widehat {T}2$ images contributes improvements to tumor segmentation of T1 images (7.89\% and 6.32\% of tumors respectively). Likewise, Table \ref{table:results_iseg2017_seg} shows that cross-modality information from $\widehat {T}1$ images leads to improvements of wm and gm segmentation of T2 images (3.08\% of wm and 2.09\% of gm). We also add cross-modality information from real modalities to make an upper bound. We observe a minor gap between results of TMS and the upper bound though our translated modalities are very close to real modalities.
It is explicable by the presence of abnormal tissue anatomy (eg. tumors) and the cortex in MR images. The tumors are diffuse and even a small difference in the overlap can cause a low value for the Dice. In addition, in some finer cortex regions (unlike large homogeneous gray matter and white matter), our approach may produce some relatively coarse images, leading to a lower Dice.
Overall, TMS outperforms the traditional FCN method when favorable cross-modality information is adopted. Fig.\ref{fig:results_segmentation} and Fig.\ref{fig:results_segmentation_seg} visualize some samples of our segmentation results on \textit{BraTs2015} and \textit{Iseg2017} respectively.

\subsection{Discussion}

We have described a new approach for cross-modality MR image generation using N2N translation network. Experimental results in section \ref{Experiments} have highlighted the capability of our proposed approach to handle complex cross-modality generation tasks. The rationales are as follows. First, the cGAN rather than GAN network is essentially conceived of as a supervised network. It not only pursues realistic looking images, but also penalizes the mismatch between input and output so as to produce grounded enough real images. Second, the L1 term, which introduces pixel-wise regularization constraints into our generation task, guarantees the quantifications of low-level textures. Besides, we also described registration and segmentation applications of generated images. Both given-modality images and generated translated-modality images are used together to provide enough contrast information to differentiate different tissues and tumors, contributing to improvements for MR images registration and segmentation. 
\begin{figure}[!ht]
\centering
\includegraphics[width=80mm ]{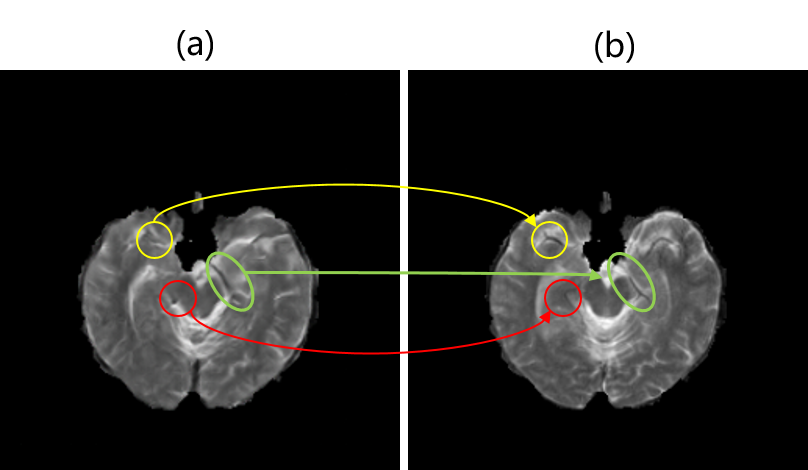}
\caption{An abortive sample in our generation results:(a) $\hat{T}2$. (b) T2. Circles in $\hat{T}2$ indicate some misdescription of tiny structures. Different colourful circles indicate different problems.}
\label{fig:discussion}
\end{figure}

Although our approach generally achieves excellent performance, we recognize that in some cases our generated images are still not as good as real images at tiny structures.
As illustrated in Fig.\ref{fig:discussion}, there are also abortive cases where tiny structures may be mistaken. In the yellow box, the eyebrow-like structure is missing. The red box indicates a non-existent round structure which might be confounded with the vessel. In the green box, the learned structure seems to be discontinuous which might give rise to perplexity for radiologists to make a diagnosis.
In the future, we will improve our algorithm to describe more tiny structures.

\section{Conclusion}
In this paper, we have developed a conditional-generative-adversarial-network-based framework for cross-modality translation that demonstrates competitive performance on cross-modality registration and segmentation. Our framework builds on top of the ideas of end-to-end NeuroImage-to-NeuroImage translation networks. We also have proposed two new approaches for MR image registration and segmentation by adopting cross-modality information from translated modality generated with our proposed framework. Our methods lead to comparable results in cross-modality generation, registration and segmentation on widely adopted MRI datasets without adding any extra data on the premise of only one modality image being given. It also suggests promising future work towards cross-modality translation tasks beyond MRI, such as from CT to MRI or from MRI to PET.
\section*{Acknowledgment}
This work is supported by Microsoft Research under the eHealth program,
the National Natural Science Foundation in China under Grant 81771910, the
National Science and Technology Major Project of the Ministry of Science
and Technology in China under Grant 2017YFC0110903, the Beijing Natural
Science Foundation in China under Grant 4152033, the Technology and
Innovation Commission of Shenzhen in China under Grant shenfagai2016-
627, Beijing Young Talent Project in China, the Fundamental Research Funds
for the Central Universities of China under Grant SKLSDE-2017ZX-08 from
the State Key Laboratory of Software Development Environment in Beihang
University in China, the 111 Project in China under Grant B13003.
The authors would like to thank all the dataset providers for making their databases publicly available.
\section*{References}
\bibliographystyle{model2-names}
\bibliography{mybibfile}

\begin{thebibliography}{62}
\expandafter\ifx\csname natexlab\endcsname\relax\def\natexlab#1{#1}\fi
\providecommand{\url}[1]{\texttt{#1}}
\providecommand{\href}[2]{#2}
\providecommand{\path}[1]{#1}
\providecommand{\DOIprefix}{doi:}
\providecommand{\ArXivprefix}{arXiv:}
\providecommand{\URLprefix}{URL: }
\providecommand{\Pubmedprefix}{pmid:}
\providecommand{\doi}[1]{\href{http://dx.doi.org/#1}{\path{#1}}}
\providecommand{\Pubmed}[1]{\href{pmid:#1}{\path{#1}}}
\providecommand{\bibinfo}[2]{#2}
\ifx\xfnm\relax \def\xfnm[#1]{\unskip,\space#1}\fi
\bibitem[{[dataset] Adriënne M.~Mendrik et~al.(2015)[dataset] Adriënne
  M.~Mendrik, Vincken, Kuijf, Breeuwer, Bouvy, Bresser, Alansary, Bruijne,
  Carass and El-Baz}]{Adri2015MRBrainS}
\bibinfo{author}{[dataset] Adriënne M.~Mendrik}, \bibinfo{author}{Vincken,
  K.L.}, \bibinfo{author}{Kuijf, H.J.}, \bibinfo{author}{Breeuwer, M.},
  \bibinfo{author}{Bouvy, W.H.}, \bibinfo{author}{Bresser, J.D.},
  \bibinfo{author}{Alansary, A.}, \bibinfo{author}{Bruijne, M.D.},
  \bibinfo{author}{Carass, A.}, \bibinfo{author}{El-Baz, A.},
  \bibinfo{year}{2015}.
\newblock \bibinfo{title}{Mrbrains challenge: Online evaluation framework for
  brain image segmentation in 3t mri scans}.
\newblock \bibinfo{journal}{Comput. Intel. Neurosc.} \bibinfo{volume}{2015},
  \bibinfo{pages}{1--16}.
\bibitem[{Artaechevarria et~al.(2009)Artaechevarria, Munoz-Barrutia and
  Ortiz-De-Solorzano}]{Artaechevarria2009Combination}
\bibinfo{author}{Artaechevarria, X.}, \bibinfo{author}{Munoz-Barrutia, A.},
  \bibinfo{author}{Ortiz-De-Solorzano, C.}, \bibinfo{year}{2009}.
\newblock \bibinfo{title}{Combination strategies in multi-atlas image
  segmentation: application to brain mr data.}
\newblock \bibinfo{journal}{IEEE Trans. Med. Imaging} \bibinfo{volume}{28},
  \bibinfo{pages}{1266--1277}.
\bibitem[{Avants et~al.(2009)Avants, Tustison and Song}]{Avants2009Advanced}
\bibinfo{author}{Avants, B.B.}, \bibinfo{author}{Tustison, N.},
  \bibinfo{author}{Song, G.}, \bibinfo{year}{2009}.
\newblock \bibinfo{title}{Advanced normalization tools (ants)}.
\newblock \bibinfo{journal}{Insight J.} \bibinfo{volume}{2},
  \bibinfo{pages}{1--35}.
\bibitem[{Balafar et~al.(2010)Balafar, Ramli, Saripan and
  Mashohor}]{Balafar2010Review}
\bibinfo{author}{Balafar, M.A.}, \bibinfo{author}{Ramli, A.R.},
  \bibinfo{author}{Saripan, M.I.}, \bibinfo{author}{Mashohor, S.},
  \bibinfo{year}{2010}.
\newblock \bibinfo{title}{Review of brain mri image segmentation methods}.
\newblock \bibinfo{journal}{Artif. Intell. Rev.} \bibinfo{volume}{33},
  \bibinfo{pages}{261--274}.
\bibitem[{Boltcheva et~al.(2009)Boltcheva, Yvinec and
  Boissonnat}]{Boltcheva2009Evaluation}
\bibinfo{author}{Boltcheva, D.}, \bibinfo{author}{Yvinec, M.},
  \bibinfo{author}{Boissonnat, J.D.}, \bibinfo{year}{2009}.
\newblock \bibinfo{title}{Evaluation of 14 nonlinear deformation algorithms
  applied to human brain mri registration.}
\newblock \bibinfo{journal}{NeuroImage} \bibinfo{volume}{46},
  \bibinfo{pages}{786--802}.
\bibitem[{Chen et~al.(2015)Chen, Li, Li, Lin, Wang, Wang, Xiao, Xu, Zhang and
  Zhang}]{Chen2015MXNet}
\bibinfo{author}{Chen, T.}, \bibinfo{author}{Li, M.}, \bibinfo{author}{Li, Y.},
  \bibinfo{author}{Lin, M.}, \bibinfo{author}{Wang, N.}, \bibinfo{author}{Wang,
  M.}, \bibinfo{author}{Xiao, T.}, \bibinfo{author}{Xu, B.},
  \bibinfo{author}{Zhang, C.}, \bibinfo{author}{Zhang, Z.},
  \bibinfo{year}{2015}.
\newblock \bibinfo{title}{Mxnet: A flexible and efficient machine learning
  library for heterogeneous distributed systems}.
\newblock \bibinfo{journal}{arXiv preprint arXiv:1512.01274} .
\bibitem[{Collobert et~al.(2011)Collobert, Kavukcuoglu and
  Farabet}]{Collobert2011Torch7}
\bibinfo{author}{Collobert, R.}, \bibinfo{author}{Kavukcuoglu, K.},
  \bibinfo{author}{Farabet, C.}, \bibinfo{year}{2011}.
\newblock \bibinfo{title}{Torch7: A matlab-like environment for machine
  learning}, in: \bibinfo{booktitle}{BigLearn, NIPS Workshop, 2011}, pp.
  \bibinfo{pages}{192376--192381}.
\bibitem[{Dou et~al.(2016)Dou, Chen, Yu, Zhao, Qin, Wang, Mok, Shi and
  Heng}]{dou2016automatic}
\bibinfo{author}{Dou, Q.}, \bibinfo{author}{Chen, H.}, \bibinfo{author}{Yu,
  L.}, \bibinfo{author}{Zhao, L.}, \bibinfo{author}{Qin, J.},
  \bibinfo{author}{Wang, D.}, \bibinfo{author}{Mok, V.C.},
  \bibinfo{author}{Shi, L.}, \bibinfo{author}{Heng, P.A.},
  \bibinfo{year}{2016}.
\newblock \bibinfo{title}{Automatic detection of cerebral microbleeds from mr
  images via 3d convolutional neural networks}.
\newblock \bibinfo{journal}{IEEE Trans. Med. Imaging} \bibinfo{volume}{35},
  \bibinfo{pages}{1182--1195}.
\bibitem[{Eugenio et~al.(2013)Eugenio, Rory and Van}]{Eugenio2013A}
\bibinfo{author}{Eugenio, I.J.}, \bibinfo{author}{Rory, S.M.},
  \bibinfo{author}{Van, L.K.}, \bibinfo{year}{2013}.
\newblock \bibinfo{title}{A unified framework for cross-modality multi-atlas
  segmentation of brain mri}.
\newblock \bibinfo{journal}{Med. Image Anal.} \bibinfo{volume}{17},
  \bibinfo{pages}{1181--1191}.
\bibitem[{Freeman and Pasztor(2000)}]{Freeman2000Learning}
\bibinfo{author}{Freeman, W.T.}, \bibinfo{author}{Pasztor, E.C.},
  \bibinfo{year}{2000}.
\newblock \bibinfo{title}{Learning low-level vision}.
\newblock \bibinfo{journal}{Int. J. Comput. Vision} \bibinfo{volume}{40},
  \bibinfo{pages}{25--47}.
\bibitem[{Goodfellow et~al.(2014)Goodfellow, Pouget-Abadie, Mirza, Xu,
  Warde-Farley, Ozair, Courville and Bengio}]{goodfellow2014generative}
\bibinfo{author}{Goodfellow, I.}, \bibinfo{author}{Pouget-Abadie, J.},
  \bibinfo{author}{Mirza, M.}, \bibinfo{author}{Xu, B.},
  \bibinfo{author}{Warde-Farley, D.}, \bibinfo{author}{Ozair, S.},
  \bibinfo{author}{Courville, A.}, \bibinfo{author}{Bengio, Y.},
  \bibinfo{year}{2014}.
\newblock \bibinfo{title}{Generative adversarial nets}, in:
  \bibinfo{booktitle}{NIPS, 2014}, pp. \bibinfo{pages}{2672--2680}.
\bibitem[{Hore and Ziou(2010)}]{Hore2010Image}
\bibinfo{author}{Hore, A.}, \bibinfo{author}{Ziou, D.}, \bibinfo{year}{2010}.
\newblock \bibinfo{title}{Image quality metrics: Psnr vs. ssim}, in:
  \bibinfo{booktitle}{ICPR}, pp. \bibinfo{pages}{2366--2369}.
\bibitem[{Huang et~al.(2017)Huang, Shao and Frangi}]{Huang2017Simultaneous}
\bibinfo{author}{Huang, Y.}, \bibinfo{author}{Shao, L.},
  \bibinfo{author}{Frangi, A.F.}, \bibinfo{year}{2017}.
\newblock \bibinfo{title}{Simultaneous super-resolution and cross-modality
  synthesis of 3d medical images using weakly-supervised joint convolutional
  sparse coding}, in: \bibinfo{booktitle}{CVPR, 2017}, pp.
  \bibinfo{pages}{5787--5796}.
\bibitem[{Iglesias et~al.(2013)Iglesias, Konukoglu, Zikic, Glocker, Leemput and
  Fischl}]{Iglesias2013Is}
\bibinfo{author}{Iglesias, J.E.}, \bibinfo{author}{Konukoglu, E.},
  \bibinfo{author}{Zikic, D.}, \bibinfo{author}{Glocker, B.},
  \bibinfo{author}{Leemput, K.V.}, \bibinfo{author}{Fischl, B.},
  \bibinfo{year}{2013}.
\newblock \bibinfo{title}{Is synthesizing mri contrast useful for
  inter-modality analysis?}, in: \bibinfo{booktitle}{MICCAI, 2013}, pp.
  \bibinfo{pages}{631--638}.
\bibitem[{Iizuka et~al.(2016)Iizuka, Simo-Serra and Ishikawa}]{iizuka2016let}
\bibinfo{author}{Iizuka, S.}, \bibinfo{author}{Simo-Serra, E.},
  \bibinfo{author}{Ishikawa, H.}, \bibinfo{year}{2016}.
\newblock \bibinfo{title}{Let there be color!: joint end-to-end learning of
  global and local image priors for automatic image colorization with
  simultaneous classification}.
\newblock \bibinfo{journal}{ACM Trans. Graph.} \bibinfo{volume}{35},
  \bibinfo{pages}{110--119}.
\bibitem[{Ioffe and Szegedy(2015)}]{Ioffe2015Batch}
\bibinfo{author}{Ioffe, S.}, \bibinfo{author}{Szegedy, C.},
  \bibinfo{year}{2015}.
\newblock \bibinfo{title}{Batch normalization: Accelerating deep network
  training by reducing internal covariate shift}, in: \bibinfo{booktitle}{ICML,
  2015}, pp. \bibinfo{pages}{448--456}.
\bibitem[{Isola et~al.(2017)Isola, Zhu, Zhou and Efros}]{isola2016image}
\bibinfo{author}{Isola, P.}, \bibinfo{author}{Zhu, J.Y.},
  \bibinfo{author}{Zhou, T.}, \bibinfo{author}{Efros, A.A.},
  \bibinfo{year}{2017}.
\newblock \bibinfo{title}{Image-to-image translation with conditional
  adversarial networks}, in: \bibinfo{booktitle}{CVPR, 2017}, pp.
  \bibinfo{pages}{5967--5976}.
\bibitem[{Jog et~al.(2013)Jog, Roy, Carass and Prince}]{Jog2013Magnetic}
\bibinfo{author}{Jog, A.}, \bibinfo{author}{Roy, S.}, \bibinfo{author}{Carass,
  A.}, \bibinfo{author}{Prince, J.L.}, \bibinfo{year}{2013}.
\newblock \bibinfo{title}{Magnetic resonance image synthesis through patch
  regression}, in: \bibinfo{booktitle}{Proc. IEEE Int. Symp. Biomed. Imaging},
  pp. \bibinfo{pages}{350--353}.
\bibitem[{Johnson et~al.(2016)Johnson, Alahi and
  Fei-Fei}]{johnson2016perceptual}
\bibinfo{author}{Johnson, J.}, \bibinfo{author}{Alahi, A.},
  \bibinfo{author}{Fei-Fei, L.}, \bibinfo{year}{2016}.
\newblock \bibinfo{title}{Perceptual losses for real-time style transfer and
  super-resolution}, in: \bibinfo{booktitle}{ECCV, 2016}, pp.
  \bibinfo{pages}{694--711}.
\bibitem[{Kingma and Ba(2014)}]{Kingma2014Adam}
\bibinfo{author}{Kingma, D.P.}, \bibinfo{author}{Ba, J.}, \bibinfo{year}{2014}.
\newblock \bibinfo{title}{Adam: A method for stochastic optimization}.
\newblock \bibinfo{journal}{arXiv preprint arXiv:1412.6980} .
\bibitem[{Klein et~al.(2010)Klein, Staring, Murphy, Viergever and
  Pluim}]{Klein2010elastix}
\bibinfo{author}{Klein, S.}, \bibinfo{author}{Staring, M.},
  \bibinfo{author}{Murphy, K.}, \bibinfo{author}{Viergever, M.A.},
  \bibinfo{author}{Pluim, J.P.}, \bibinfo{year}{2010}.
\newblock \bibinfo{title}{elastix: a toolbox for intensity-based medical image
  registration.}
\newblock \bibinfo{journal}{IEEE Trans. Med. Imaging} \bibinfo{volume}{29},
  \bibinfo{pages}{196--205}.
\bibitem[{Larsen et~al.(2015)Larsen, Sønderby, Larochelle and
  Winther}]{Larsen2015Autoencoding}
\bibinfo{author}{Larsen, A.B.L.}, \bibinfo{author}{Sønderby, S.K.},
  \bibinfo{author}{Larochelle, H.}, \bibinfo{author}{Winther, O.},
  \bibinfo{year}{2015}.
\newblock \bibinfo{title}{Autoencoding beyond pixels using a learned similarity
  metric}.
\newblock \bibinfo{journal}{arXiv preprint arXiv:1512.09300} ,
  \bibinfo{pages}{1558--1566}.
\bibitem[{Larsson et~al.(2016)Larsson, Maire and
  Shakhnarovich}]{larsson2016learning}
\bibinfo{author}{Larsson, G.}, \bibinfo{author}{Maire, M.},
  \bibinfo{author}{Shakhnarovich, G.}, \bibinfo{year}{2016}.
\newblock \bibinfo{title}{Learning representations for automatic colorization},
  in: \bibinfo{booktitle}{ECCV, 2016}, pp. \bibinfo{pages}{577--593}.
\bibitem[{Lazarow et~al.(2017a)Lazarow, Jin and Tu}]{Lazarow2017Introspective}
\bibinfo{author}{Lazarow, J.}, \bibinfo{author}{Jin, L.}, \bibinfo{author}{Tu,
  Z.}, \bibinfo{year}{2017}a.
\newblock \bibinfo{title}{Introspective neural networks for generative
  modeling}.
\newblock \bibinfo{journal}{ICCV, 2017} , \bibinfo{pages}{5907--5915}.
\bibitem[{Lazarow et~al.(2017b)Lazarow, Jin and Tu}]{Jin2017Introspective}
\bibinfo{author}{Lazarow, J.}, \bibinfo{author}{Jin, L.}, \bibinfo{author}{Tu,
  Z.}, \bibinfo{year}{2017}b.
\newblock \bibinfo{title}{Introspective neural networks for generative
  modeling}, in: \bibinfo{booktitle}{CVPR, 2017}, pp.
  \bibinfo{pages}{2774--2783}.
\bibitem[{Lee et~al.(2014)Lee, Xie, Gallagher, Zhang and Tu}]{Lee2014Deeply}
\bibinfo{author}{Lee, C.Y.}, \bibinfo{author}{Xie, S.},
  \bibinfo{author}{Gallagher, P.}, \bibinfo{author}{Zhang, Z.},
  \bibinfo{author}{Tu, Z.}, \bibinfo{year}{2014}.
\newblock \bibinfo{title}{Deeply-supervised nets}.
\newblock \bibinfo{journal}{Artif. Intell.} , \bibinfo{pages}{562--570}.
\bibitem[{Long et~al.(2015)Long, Shelhamer and Darrell}]{long2015fully}
\bibinfo{author}{Long, J.}, \bibinfo{author}{Shelhamer, E.},
  \bibinfo{author}{Darrell, T.}, \bibinfo{year}{2015}.
\newblock \bibinfo{title}{Fully convolutional networks for semantic
  segmentation}, in: \bibinfo{booktitle}{CVPR, 2015}, pp.
  \bibinfo{pages}{3431--3440}.
\bibitem[{[dataset] Menze et~al.(2015)[dataset] Menze, Jakab, Bauer,
  Kalpathy-Cramer, Farahani, Kirby, Burren, Porz, Slotboom and
  Wiest}]{Menze2015The}
\bibinfo{author}{[dataset] Menze, B.H.}, \bibinfo{author}{Jakab, A.},
  \bibinfo{author}{Bauer, S.}, \bibinfo{author}{Kalpathy-Cramer, J.},
  \bibinfo{author}{Farahani, K.}, \bibinfo{author}{Kirby, J.},
  \bibinfo{author}{Burren, Y.}, \bibinfo{author}{Porz, N.},
  \bibinfo{author}{Slotboom, J.}, \bibinfo{author}{Wiest, R.},
  \bibinfo{year}{2015}.
\newblock \bibinfo{title}{The multimodal brain tumor image segmentation
  benchmark (brats)}.
\newblock \bibinfo{journal}{IEEE Trans. Med. Imaging} \bibinfo{volume}{34},
  \bibinfo{pages}{1993--2024}.
\bibitem[{Miller et~al.(1993)Miller, Christensen, Amit and
  Grenander}]{Miller1993Mathematical}
\bibinfo{author}{Miller, M.I.}, \bibinfo{author}{Christensen, G.E.},
  \bibinfo{author}{Amit, Y.}, \bibinfo{author}{Grenander, U.},
  \bibinfo{year}{1993}.
\newblock \bibinfo{title}{Mathematical textbook of deformable neuroanatomies.}
\newblock \bibinfo{journal}{Proc. Acad. Nat. Sci. Phila.} \bibinfo{volume}{90},
  \bibinfo{pages}{11944--11948}.
\bibitem[{Mirza and Osindero(2014)}]{mirza2014conditional}
\bibinfo{author}{Mirza, M.}, \bibinfo{author}{Osindero, S.},
  \bibinfo{year}{2014}.
\newblock \bibinfo{title}{Conditional generative adversarial nets}, in:
  \bibinfo{booktitle}{ICLR, 2014}, pp. \bibinfo{pages}{2672--2680}.
\bibitem[{Nie et~al.(2017)Nie, Trullo, Petitjean, Ruan and
  Shen}]{nie2016medical}
\bibinfo{author}{Nie, D.}, \bibinfo{author}{Trullo, R.},
  \bibinfo{author}{Petitjean, C.}, \bibinfo{author}{Ruan, S.},
  \bibinfo{author}{Shen, D.}, \bibinfo{year}{2017}.
\newblock \bibinfo{title}{Medical image synthesis with context-aware generative
  adversarial networks}, in: \bibinfo{booktitle}{MICCAI, 2017}, pp.
  \bibinfo{pages}{417--425}.
\bibitem[{Pathak et~al.(2016)Pathak, Krahenbuhl, Donahue, Darrell and
  Efros}]{pathak2016context}
\bibinfo{author}{Pathak, D.}, \bibinfo{author}{Krahenbuhl, P.},
  \bibinfo{author}{Donahue, J.}, \bibinfo{author}{Darrell, T.},
  \bibinfo{author}{Efros, A.A.}, \bibinfo{year}{2016}.
\newblock \bibinfo{title}{Context encoders: Feature learning by inpainting},
  in: \bibinfo{booktitle}{CVPR, 2016}, pp. \bibinfo{pages}{2536--2544}.
\bibitem[{Pedregosa et~al.(2011)Pedregosa, Varoquaux, Gramfort, Michel,
  Thirion, Grisel, Blondel, Prettenhofer, Weiss, Dubourg, Vanderplas, Passos,
  Cournapeau, Brucher, Perrot and Duchesnay}]{scikit-learn}
\bibinfo{author}{Pedregosa, F.}, \bibinfo{author}{Varoquaux, G.},
  \bibinfo{author}{Gramfort, A.}, \bibinfo{author}{Michel, V.},
  \bibinfo{author}{Thirion, B.}, \bibinfo{author}{Grisel, O.},
  \bibinfo{author}{Blondel, M.}, \bibinfo{author}{Prettenhofer, P.},
  \bibinfo{author}{Weiss, R.}, \bibinfo{author}{Dubourg, V.},
  \bibinfo{author}{Vanderplas, J.}, \bibinfo{author}{Passos, A.},
  \bibinfo{author}{Cournapeau, D.}, \bibinfo{author}{Brucher, M.},
  \bibinfo{author}{Perrot, M.}, \bibinfo{author}{Duchesnay, E.},
  \bibinfo{year}{2011}.
\newblock \bibinfo{title}{Scikit-learn: Machine learning in {P}ython}.
\newblock \bibinfo{journal}{J. Mach. Learn Res.} \bibinfo{volume}{12},
  \bibinfo{pages}{2825--2830}.
\bibitem[{Penney et~al.(1998)Penney, Weese, Little, Desmedt, Hill and
  Hawkes}]{Penney1998A}
\bibinfo{author}{Penney, G.P.}, \bibinfo{author}{Weese, J.},
  \bibinfo{author}{Little, J.A.}, \bibinfo{author}{Desmedt, P.},
  \bibinfo{author}{Hill, D.L.G.}, \bibinfo{author}{Hawkes, D.J.},
  \bibinfo{year}{1998}.
\newblock \bibinfo{title}{A comparison of similarity measures for use in
  2-d-3-d medical image registration}.
\newblock \bibinfo{journal}{IEEE Trans. Med. Imaging} \bibinfo{volume}{17},
  \bibinfo{pages}{586--595}.
\bibitem[{Pinheiro and Collobert(2015)}]{Pinheiro2015From}
\bibinfo{author}{Pinheiro, P.O.}, \bibinfo{author}{Collobert, R.},
  \bibinfo{year}{2015}.
\newblock \bibinfo{title}{From image-level to pixel-level labeling with
  convolutional networks}.
\newblock \bibinfo{journal}{CVPR, 2015} , \bibinfo{pages}{1713--1721}.
\bibitem[{Pluim et~al.(2003)Pluim, Maintz and Viergever}]{Pluim2003Mutual}
\bibinfo{author}{Pluim, J.P.W.}, \bibinfo{author}{Maintz, J.B.A.},
  \bibinfo{author}{Viergever, M.A.}, \bibinfo{year}{2003}.
\newblock \bibinfo{title}{Mutual-information-based registration of medical
  images: a survey}.
\newblock \bibinfo{journal}{IEEE Trans. Med. Imaging} \bibinfo{volume}{22},
  \bibinfo{pages}{986--1004}.
\bibitem[{Radford et~al.(2015)Radford, Metz and
  Chintala}]{radford2015unsupervised}
\bibinfo{author}{Radford, A.}, \bibinfo{author}{Metz, L.},
  \bibinfo{author}{Chintala, S.}, \bibinfo{year}{2015}.
\newblock \bibinfo{title}{Unsupervised representation learning with deep
  convolutional generative adversarial networks}.
\newblock \bibinfo{journal}{arXiv preprint arXiv:1511.06434} .
\bibitem[{Ronneberger et~al.(2015)Ronneberger, Fischer and
  Brox}]{ronneberger2015u}
\bibinfo{author}{Ronneberger, O.}, \bibinfo{author}{Fischer, P.},
  \bibinfo{author}{Brox, T.}, \bibinfo{year}{2015}.
\newblock \bibinfo{title}{U-net: Convolutional networks for biomedical image
  segmentation}, in: \bibinfo{booktitle}{MICCAI, 2015}, pp.
  \bibinfo{pages}{234--241}.
\bibitem[{Rousseau(2008)}]{Rousseau2008Brain}
\bibinfo{author}{Rousseau, F.}, \bibinfo{year}{2008}.
\newblock \bibinfo{title}{Brain hallucination}, in: \bibinfo{booktitle}{ECCV,
  2008}, pp. \bibinfo{pages}{497--508}.
\bibitem[{Roy et~al.(2013)Roy, Carass and Prince}]{Roy2013Magnetic}
\bibinfo{author}{Roy, S.}, \bibinfo{author}{Carass, A.},
  \bibinfo{author}{Prince, J.}, \bibinfo{year}{2013}.
\newblock \bibinfo{title}{Magnetic resonance image example based contrast
  synthesis.}
\newblock \bibinfo{journal}{IEEE Trans. Med. Imaging} \bibinfo{volume}{32},
  \bibinfo{pages}{2348--2363}.
\bibitem[{Rueckert et~al.(1999)Rueckert, Sonoda, Hayes, Hill, Leach and
  Hawkes}]{Rueckert1999Nonrigid}
\bibinfo{author}{Rueckert, D.}, \bibinfo{author}{Sonoda, L.I.},
  \bibinfo{author}{Hayes, C.}, \bibinfo{author}{Hill, D.L.G.},
  \bibinfo{author}{Leach, M.O.}, \bibinfo{author}{Hawkes, D.J.},
  \bibinfo{year}{1999}.
\newblock \bibinfo{title}{Nonrigid registration using free-form deformations:
  application to breast mr images}.
\newblock \bibinfo{journal}{IEEE Trans. Med. Imaging} \bibinfo{volume}{18},
  \bibinfo{pages}{712--721}.
\bibitem[{Rzedzian et~al.(1983)Rzedzian, Chapman, Mansfield, Coupland, Doyle,
  Chrispin, Guilfoyle and Small}]{Rzedzian1983Real}
\bibinfo{author}{Rzedzian, R.}, \bibinfo{author}{Chapman, B.},
  \bibinfo{author}{Mansfield, P.}, \bibinfo{author}{Coupland, R.E.},
  \bibinfo{author}{Doyle, M.}, \bibinfo{author}{Chrispin, A.},
  \bibinfo{author}{Guilfoyle, D.}, \bibinfo{author}{Small, P.},
  \bibinfo{year}{1983}.
\newblock \bibinfo{title}{Real-time nuclear magnetic resonance clinical imaging
  in paediatrics}.
\newblock \bibinfo{journal}{Lancet} \bibinfo{volume}{2},
  \bibinfo{pages}{1281--1282}.
\bibitem[{Sasirekha and Kashwan(2015)}]{Sasirekha2015Improved}
\bibinfo{author}{Sasirekha, N.}, \bibinfo{author}{Kashwan, K.},
  \bibinfo{year}{2015}.
\newblock \bibinfo{title}{Improved segmentation of mri brain images by
  denoising and contrast enhancement}.
\newblock \bibinfo{journal}{Indian J. Sci. Technol.} \bibinfo{volume}{8},
  \bibinfo{pages}{1--7}.
\bibitem[{Srivastava(2013)}]{srivastava2013improving}
\bibinfo{author}{Srivastava, N.}, \bibinfo{year}{2013}.
\newblock \bibinfo{title}{Improving neural networks with dropout}.
\newblock \bibinfo{journal}{UofT} \bibinfo{volume}{182}, \bibinfo{pages}{566}.
\bibitem[{Tsao(2010)}]{Tsao2010Ultrafast}
\bibinfo{author}{Tsao, J.}, \bibinfo{year}{2010}.
\newblock \bibinfo{title}{Ultrafast imaging: principles, pitfalls, solutions,
  and applications}.
\newblock \bibinfo{journal}{J. Magn. Reson. Imag.} \bibinfo{volume}{32},
  \bibinfo{pages}{252--266}.
\bibitem[{Tseng et~al.(2017)Tseng, Lin, Hsu and Huang}]{Tseng2017Joint}
\bibinfo{author}{Tseng, K.L.}, \bibinfo{author}{Lin, Y.L.},
  \bibinfo{author}{Hsu, W.}, \bibinfo{author}{Huang, C.Y.},
  \bibinfo{year}{2017}.
\newblock \bibinfo{title}{Joint sequence learning and cross-modality
  convolution for 3d biomedical segmentation}, in: \bibinfo{booktitle}{CVPR,
  2017}, pp. \bibinfo{pages}{3739--3746}.
\bibitem[{Tu(2007)}]{Tu2007Learning}
\bibinfo{author}{Tu, Z.}, \bibinfo{year}{2007}.
\newblock \bibinfo{title}{Learning generative models via discriminative
  approaches}, in: \bibinfo{booktitle}{CVPR, 2007}, pp. \bibinfo{pages}{1--8}.
\bibitem[{Ulyanov et~al.(2016)Ulyanov, Vedaldi and
  Lempitsky}]{Ulyanov2016Instance}
\bibinfo{author}{Ulyanov, D.}, \bibinfo{author}{Vedaldi, A.},
  \bibinfo{author}{Lempitsky, V.}, \bibinfo{year}{2016}.
\newblock \bibinfo{title}{Instance normalization: The missing ingredient for
  fast stylization}.
\newblock \bibinfo{journal}{arXiv preprint arXiv:1607.08022} .
\bibitem[{Van~Nguyen et~al.(2015)Van~Nguyen, Zhou and
  Vemulapalli}]{Nguyen2015Cross}
\bibinfo{author}{Van~Nguyen, H.}, \bibinfo{author}{Zhou, K.},
  \bibinfo{author}{Vemulapalli, R.}, \bibinfo{year}{2015}.
\newblock \bibinfo{title}{Cross-domain synthesis of medical images using
  efficient location-sensitive deep network}, in: \bibinfo{booktitle}{MICCAI,
  2015}, pp. \bibinfo{pages}{677--684}.
\bibitem[{Vemulapalli et~al.(2016)Vemulapalli, Nguyen and
  Zhou}]{Vemulapalli2016Unsupervised}
\bibinfo{author}{Vemulapalli, R.}, \bibinfo{author}{Nguyen, H.V.},
  \bibinfo{author}{Zhou, S.K.}, \bibinfo{year}{2016}.
\newblock \bibinfo{title}{Unsupervised cross-modal synthesis of
  subject-specific scans}, in: \bibinfo{booktitle}{ICCV, 2016}, pp.
  \bibinfo{pages}{630--638}.
\bibitem[{Viola and Wells(1997)}]{Viola1997Alignment}
\bibinfo{author}{Viola, P.}, \bibinfo{author}{Wells, W.}, \bibinfo{year}{1997}.
\newblock \bibinfo{title}{Alignment by maximization of mutual information}.
\newblock \bibinfo{journal}{Int. J. Comput. Vision} \bibinfo{volume}{24},
  \bibinfo{pages}{137--154}.
\bibitem[{Wang et~al.(2013)Wang, Suh, Das, Pluta, Craige and
  Yushkevich}]{Wang2013Multi}
\bibinfo{author}{Wang, H.}, \bibinfo{author}{Suh, J.W.}, \bibinfo{author}{Das,
  S.R.}, \bibinfo{author}{Pluta, J.B.}, \bibinfo{author}{Craige, C.},
  \bibinfo{author}{Yushkevich, P.A.}, \bibinfo{year}{2013}.
\newblock \bibinfo{title}{Multi-atlas segmentation with joint label fusion}.
\newblock \bibinfo{journal}{IEEE Trans. Pattern Anal. Mach. Intel}
  \bibinfo{volume}{35}, \bibinfo{pages}{611--623}.
\bibitem[{[dataset] Wang et~al.(2015)[dataset] Wang, Gao, Shi, Li, Gilmore, Lin
  and Shen}]{wang2015links}
\bibinfo{author}{[dataset] Wang, L.}, \bibinfo{author}{Gao, Y.},
  \bibinfo{author}{Shi, F.}, \bibinfo{author}{Li, G.},
  \bibinfo{author}{Gilmore, J.H.}, \bibinfo{author}{Lin, W.},
  \bibinfo{author}{Shen, D.}, \bibinfo{year}{2015}.
\newblock \bibinfo{title}{Links: Learning-based multi-source integration
  framework for segmentation of infant brain images}.
\newblock \bibinfo{journal}{NeuroImage} \bibinfo{volume}{108},
  \bibinfo{pages}{160--172}.
\bibitem[{Wang and Gupta(2016)}]{wang2016generative}
\bibinfo{author}{Wang, X.}, \bibinfo{author}{Gupta, A.}, \bibinfo{year}{2016}.
\newblock \bibinfo{title}{Generative image modeling using style and structure
  adversarial networks}, in: \bibinfo{booktitle}{ECCV, 2016}, pp.
  \bibinfo{pages}{318--335}.
\bibitem[{Wang and Bovik(2009)}]{Wang2009Mean}
\bibinfo{author}{Wang, Z.}, \bibinfo{author}{Bovik, A.C.},
  \bibinfo{year}{2009}.
\newblock \bibinfo{title}{Mean squared error: Love it or leave it? a new look
  at signal fidelity measures}.
\newblock \bibinfo{journal}{IEEE Signal Process. Mag.} \bibinfo{volume}{26},
  \bibinfo{pages}{98--117}.
\bibitem[{West et~al.(1997)West, Fitzpatrick, Wang, Dawant, Jr, Kessler,
  Maciunas, Barillot, Lemoine and Collignon}]{West1997Comparison}
\bibinfo{author}{West, J.}, \bibinfo{author}{Fitzpatrick, J.M.},
  \bibinfo{author}{Wang, M.Y.}, \bibinfo{author}{Dawant, B.M.},
  \bibinfo{author}{Jr, M.C.}, \bibinfo{author}{Kessler, R.M.},
  \bibinfo{author}{Maciunas, R.J.}, \bibinfo{author}{Barillot, C.},
  \bibinfo{author}{Lemoine, D.}, \bibinfo{author}{Collignon, A.},
  \bibinfo{year}{1997}.
\newblock \bibinfo{title}{Comparison and evaluation of retrospective
  intermodality brain image registration techniques}.
\newblock \bibinfo{journal}{J. Cmoput. Assist. Tomo.} \bibinfo{volume}{21},
  \bibinfo{pages}{554--566}.
\bibitem[{Wolterink et~al.(2017)Wolterink, Leiner, Viergever and
  Isgum}]{WolterinkGenerative}
\bibinfo{author}{Wolterink, J.M.}, \bibinfo{author}{Leiner, T.},
  \bibinfo{author}{Viergever, M.A.}, \bibinfo{author}{Isgum, I.},
  \bibinfo{year}{2017}.
\newblock \bibinfo{title}{Generative adversarial networks for noise reduction
  in low-dose ct}.
\newblock \bibinfo{journal}{IEEE Trans. Med. Imaging} \bibinfo{volume}{36},
  \bibinfo{pages}{2536--2545}.
\bibitem[{Xie and Tu(2015)}]{Xie2015Holistically}
\bibinfo{author}{Xie, S.}, \bibinfo{author}{Tu, Z.}, \bibinfo{year}{2015}.
\newblock \bibinfo{title}{Holistically-nested edge detection}.
\newblock \bibinfo{journal}{ICCV, 2015} , \bibinfo{pages}{1--16}.
\bibitem[{Xu et~al.(2017)Xu, Li, Wang, Liu, Fan, Lai and Chang}]{xu2017gland}
\bibinfo{author}{Xu, Y.}, \bibinfo{author}{Li, Y.}, \bibinfo{author}{Wang, Y.},
  \bibinfo{author}{Liu, M.}, \bibinfo{author}{Fan, Y.}, \bibinfo{author}{Lai,
  M.}, \bibinfo{author}{Chang, E.}, \bibinfo{year}{2017}.
\newblock \bibinfo{title}{Gland instance segmentation using deep multichannel
  neural networks}.
\newblock \bibinfo{journal}{IEEE Trans. Biomed. Eng.} \bibinfo{volume}{64},
  \bibinfo{pages}{2901--2912}.
\bibitem[{Yoo et~al.(2016)Yoo, Kim, Park, Paek and Kweon}]{yoo2016pixel}
\bibinfo{author}{Yoo, D.}, \bibinfo{author}{Kim, N.}, \bibinfo{author}{Park,
  S.}, \bibinfo{author}{Paek, A.S.}, \bibinfo{author}{Kweon, I.S.},
  \bibinfo{year}{2016}.
\newblock \bibinfo{title}{Pixel-level domain transfer}, in:
  \bibinfo{booktitle}{ECCV, 2016}, pp. \bibinfo{pages}{517--532}.
\bibitem[{Zhang et~al.(2017)Zhang, Xu, Li, Zhang, Huang, Wang and
  Metaxas}]{zhang2016stackgan}
\bibinfo{author}{Zhang, H.}, \bibinfo{author}{Xu, T.}, \bibinfo{author}{Li,
  H.}, \bibinfo{author}{Zhang, S.}, \bibinfo{author}{Huang, X.},
  \bibinfo{author}{Wang, X.}, \bibinfo{author}{Metaxas, D.},
  \bibinfo{year}{2017}.
\newblock \bibinfo{title}{Stackgan: Text to photo-realistic image synthesis
  with stacked generative adversarial networks}, in: \bibinfo{booktitle}{ICCV,
  2017}, pp. \bibinfo{pages}{5907--5915}.
\bibitem[{Zhou and Berg(2016)}]{zhou2016learning}
\bibinfo{author}{Zhou, Y.}, \bibinfo{author}{Berg, T.L.}, \bibinfo{year}{2016}.
\newblock \bibinfo{title}{Learning temporal transformations from time-lapse
  videos}, in: \bibinfo{booktitle}{ECCV, 2016}, pp. \bibinfo{pages}{262--277}.

\end{thebibliography}
\end{document}